\documentclass[10pt,twocolumn, review, letterpaper]{article}

\newcommand{\final}{1}
\newcommand{\forReview}{0}
\newcommand{\forArxiv}{1}

\usepackage{iccv}
\usepackage{times}
\usepackage{epsfig}
\usepackage{graphicx}
\usepackage{amsmath}
\usepackage{amssymb}
\usepackage{amsthm}
\usepackage{subcaption}
\usepackage{gensymb}
\usepackage{authblk}
\ifdefined\siggraph
\usepackage{times}
\fi

\usepackage{color}
\usepackage{ifthen}
\usepackage{float}
\usepackage{alltt}
\usepackage{newlfont} 
\usepackage{wrapfig}
\usepackage{booktabs}
\usepackage{multirow}
\usepackage{array}
\usepackage{appendix}

\usepackage{comment}
\usepackage{amsmath}
\usepackage{amssymb}
\usepackage{amsthm}
\usepackage{textcomp}

\definecolor{DeltaColor}{rgb}{0.039,0.73,0.71}
\definecolor{SetaColor}{rgb}{0.867, 0.0235, 0.376}
\definecolor{SigmaColor}{rgb}{0.98,0.45,0.0}
\definecolor{RedColor}{rgb}{0.8,0,0}
\definecolor{AlphaColor}{rgb}{0,0,0.8}
\definecolor{BetaColor}{rgb}{0.8,0,0.8}
\definecolor{GammaColor}{rgb}{0.5,0,0.7}
\definecolor{EpsilonColor}{rgb}{0.353,0.725,0.906}
\definecolor{TauColor}{rgb}{0.423,0.235,0.192}
\newcommand{\weikai}[1]{{\color{RedColor} Weikai: #1 $\qed$}}
\newcommand{\shichen}[1]{{\color{AlphaColor} Shichen: #1 $\qed$}}
\newcommand{\tianye}[1]{{\color{SigmaColor} Tianye: #1 $\qed$}}
\newcommand{\jun}[1]{{\color{GammaColor} Jun: #1 $\qed$}}
\newcommand{\yajie}[1]{{\color{DeltaColor} Yajie: #1 $\qed$}}
\newcommand{\hao}[1]{{\color{HaoColor} Hao: #1 $\qed$}}
\newcommand{\andrew}[1]{{\color{EpsilonColor} Andrew: #1 $\qed$}}
\newcommand{\paul}[1]{{\color{TauColor} Paul: #1 $\qed$}}

\newcommand{\warning}[1]{{\it\color{red} #1}}
\newcommand{\note}[1]{{\it\color{blue} #1}}
\newcommand{\nothing}[1]{}

\definecolor{AudioColor}{rgb}{0.56,0.34,0.62}

\definecolor{DeadlineColor}{rgb}{0.9,0.4,0} 
\newcommand{\deadline}[1]{{\bf\color{DeadlineColor} ETA: #1}}

\definecolor{figred}{rgb}{1,0,0}
\definecolor{figgreen}{rgb}{0,0.6,0}
\definecolor{figblue}{rgb}{0,0,1}
\definecolor{figpink}{rgb}{1,0.63,0.63}

\newcolumntype{C}[1]{>{\centering}m{#1}}

\ifthenelse{\equal{\final}{1}}
{
\renewcommand{\shichen}[1]{}
\renewcommand{\weikai}[1]{}
\renewcommand{\tianye}[1]{}
\renewcommand{\jun}[1]{}
\renewcommand{\yajie}[1]{}
\renewcommand{\andrew}[1]{}
\renewcommand{\hao}[1]{}
\renewcommand{\paul}[1]{}
\renewcommand{\warning}[1]{}
\renewcommand{\note}[1]{}
\renewcommand{\deadline}[1]{}
}
{}

\newcounter{pccount}
\floatstyle{plain}

\newcommand{\filename}[1]{\url{#1}}
\newcommand{\foldername}[1]{\url{#1}}

\newcommand{\argminE}{\mathop{\mathrm{argmin}}}  
\DeclareMathOperator*{\argmin}{argmin}         

\newcommand{\model}{soft rasterizer}

\newcommand{\modelshort}{SoftRas }

\newcommand{\aggregate}{\mathcal{A}}
\newcommand{\Dmap}{\mathcal{D}}
\newcommand{\D}{D}
\newcommand{\pixel}{p}
\newcommand{\face}{f}
\newcommand{\dist}{d}

\newcommand{\sil}{\hat{I_s}}
\newcommand{\silGT}{I_s}
\newcommand{\rgb}{\hat{I_c}}
\newcommand{\rgbGT}{I_c}

\newcommand{\img}{\mathbf{I}}
\newcommand{\loss}{\mathcal{L}}

\newcommand{\sig}{\delta}

\newcommand{\geometry}{\mathbf{M}}
\newcommand{\appearance}{\mathbf{A}}
\newcommand{\camera}{\mathbf{P}}
\newcommand{\lighting}{\mathbf{L}}
\newcommand{\normal}{\mathbf{N}}
\newcommand{\depth}{\mathbf{Z}}
\newcommand{\uv}{\mathbf{U}} 
\newcommand{\fragment}{\mathbf{F}}
\newcommand{\vcolor}{\mathbf{C}}

\newcommand{\weight}{w}
\newcommand{\backColor}{C_b}

\newcommand{\equFace}{\mathbf{U}}
\newcommand{\equPixel}{\mathbf{p}}
\newcommand{\equX}{x}
\newcommand{\equY}{y}
\newcommand{\equDis}{\mathcal{D}}

\newcommand{\equDMetric}{D}
\newcommand{\equSig}{\delta}
\newcommand{\equZ}{z}
\newcommand{\equZNear}{Z_{near}}
\newcommand{\equZFar}{Z_{far}}
\newcommand{\equColor}{C}
\newcommand{\equImage}{I}

\newcommand{\equBaryMat}{\mathbf{b}}
\newcommand{\equBaryTMat}{\mathbf{t}}
\newcommand{\equGamma}{\gamma}
\newcommand{\equSigma}{\sigma}

\newcommand{\equSum}{w}
\newcommand{\equPartial}[2]{\frac{\partial #1}{\partial #2}}

\ifthenelse{\equal{\forReview}{1}}
{
	\usepackage[pagebackref=true,breaklinks=true,letterpaper=true,colorlinks,bookmarks=false]{hyperref}
	
	\def\iccvPaperID{*****} 

	\ificcvfinal\pagestyle{empty}\fi
}
{
	\usepackage[pagebackref=true,breaklinks=true,colorlinks,bookmarks=false]{hyperref}

	\iccvfinalcopy 
	
	\def\iccvPaperID{****} 

	\setcounter{page}{1}
}

\graphicspath{
	{figs/raster/}
	{figs/handdrawn/}
}

\begin{document}
	
\linespread{0.9}

\title{Soft Rasterizer: A Differentiable Renderer 
	for Image-based 3D Reasoning}


\author[1,2]{Shichen Liu}
\author[1,2]{Tianye Li}
\author[1]{Weikai Chen}
\author[1,2,3]{Hao Li}
\affil[1]{USC Institute for Creative Technologies}
\affil[2]{University of Southern California}
\affil[3]{Pinscreen}
\affil[ ]{{\tt\small\{\href{mailto:lshichen@ict.usc.edu}{lshichen},
\href{mailto:tli@ict.usc.edu}{tli}, \href{mailto:wechen@ict.usc.edu}{wechen}\}@ict.usc.edu \quad \href{mailto:hao@hao-li.com}{hao@hao-li.com}}}

\maketitle
\nothing{
\note{
	Summary of Intro (make short and highlight the most important messages)
	\begin{itemize}
		\item What s the problem?
		Inverse graphics is key to reconstruct 3D shape from images. However, rendering is not differentiable due to the discrete rasterization operation. 
		\item What do we introduce?
		\item How does it roughly work? (main differentiator to existing methods? What's unique?)
		\item What can we do now?
		\item What do we show?
		\item What's the impact?
	\end{itemize}
}
}

\begin{abstract}
Rendering bridges the gap between 2D vision and 3D scenes by simulating the physical process of image formation. By inverting such renderer, one can think of a learning approach to infer 3D information from 2D images. However, standard graphics renderers involve a fundamental discretization step called rasterization, which prevents the rendering process to be differentiable, hence able to be learned. Unlike the state-of-the-art differentiable renderers \cite{loper2014opendr,kato2018neural}, which only approximate the rendering gradient in the back propagation, we propose a truly differentiable rendering framework that is able to (1) directly render colorized mesh using differentiable functions and (2) back-propagate efficient supervision signals to mesh vertices and their attributes from various forms of image representations, including silhouette, shading and color images. The key to our framework is a novel formulation that views rendering as an aggregation function that fuses the probabilistic contributions of all mesh triangles with respect to the rendered pixels. Such formulation enables our framework to flow gradients to the occluded and far-range vertices, which cannot be achieved by the previous state-of-the-arts. We show that by using the proposed renderer, one can achieve significant improvement in 3D unsupervised single-view reconstruction both qualitatively and quantitatively. Experiments also demonstrate that our approach is able to handle the challenging tasks in image-based shape fitting, which remain nontrivial to existing differentiable renderers. 
Code is available at \url{https://github.com/ShichenLiu/SoftRas}.


\end{abstract}

\vspace{-2mm}
\section{Introduction}
\label{sec:intro}

\nothing{
\weikai{Our advantages:
	\begin{itemize}
		\item Unbiased estimation of rasterization. Enable training in both forward and backward directions.
		\item Unsupervised training without the need of 3D groundtruth. Reduce a huge amount of labeling effort. 
		\item Novel approach of color reconstruction without use of the texture map. Can be applied to arbitrary genus-0 surface.
	\end{itemize}
}	
}

\nothing{
\note{Why differentiable renderer is important?
\begin{itemize}
	\item Image based 3D reconstruction problem is important
	\item Supervised learning requires huge amount of 3D ground truth, which is difficult to obtain. The key to unsupervised learning is differentiable rendering
\end{itemize}	
}
}

\begin{figure}[h!]
  	\centering
	\includegraphics[width=1\linewidth]{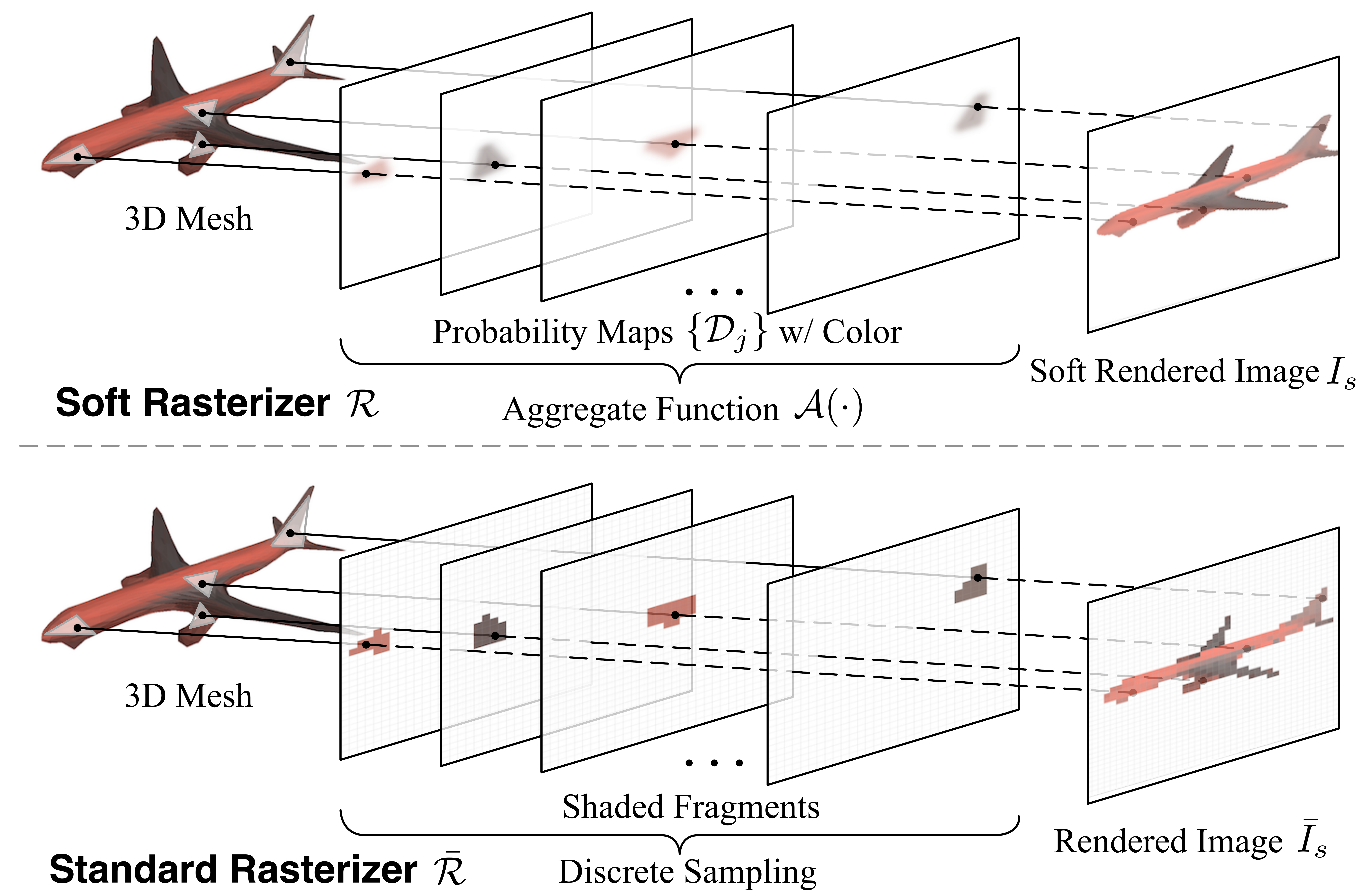}
\caption{We propose Soft Rasterizer $\mathcal{R}$ (upper), a truly differentiable renderer, which formulates rendering as a differentiable aggregating process $\aggregate(\cdot)$ that fuses per-triangle contributions $\{\Dmap_i\}$ in a ``soft" probabilistic manner.
Our approach attacks the core problem of differentiating the standard rasterizer, which cannot flow gradients from pixels to geometry due to the discrete sampling operation (below).  
}
  	\label{fig:teaser}
\end{figure}

\nothing{We propose the differentiable Soft Rasterizer (upper) to approximate the standard rasterizer (below) in graphics pipeline. We decompose rasterization into: 1) triangle-wise regions $\{\overline{D}_i\}$ in camera coordinate; 2) aggregate function $A(\cdot)$ that takes all regions to produce the raster image. In this perspective, we soften the hard triangle regions with signed distance maps $\{D_i\}$ so that the whole process will be differentiable. The proposed rasterizer could be integrated into neural networks.}

\begin{figure*}
	\centering

	\includegraphics[width=1.0\linewidth]{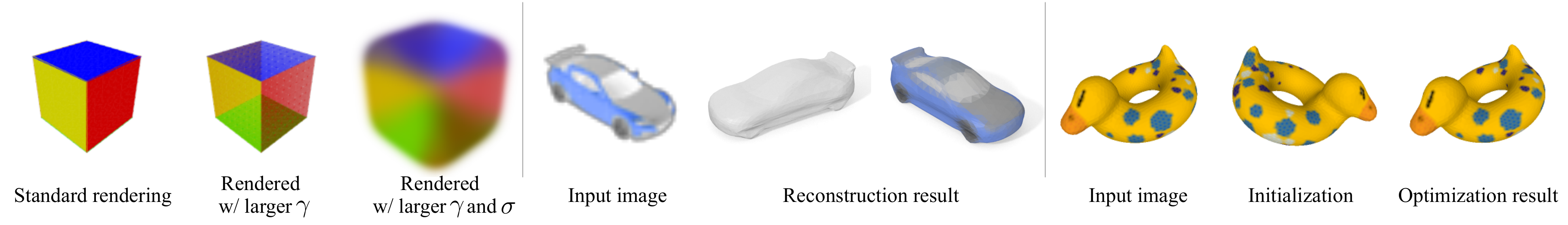}

	\vspace{-6pt}
	\caption{
	Forward rendering: various rendering effects generated by SoftRas (left). Different degrees of transparency and blurriness can be achieved by tuning $\gamma$ and $\sigma$ respectively.
	Applications based on the backward gradients provided by SoftRas: (1) 3D unsupervised mesh reconstruction from a single input image (middle) and (2) 3D pose fitting to the target image by flowing gradient to the occluded triangles (right). 
		}
	\label{fig:teaser2}
	\vspace{-2mm}
\end{figure*}


Understanding and reconstructing 3D scenes and structures from 2D images has been one of the fundamental goals in computer vision.
The key to image-based 3D reasoning is to find sufficient supervisions flowing from the pixels to the 3D properties. 
To obtain image-to-3D correlations, prior approaches mainly rely on the matching losses based on 2D key points/contours~\cite{bogo2016keep, pavlakos2017coarse,liu2016joint,matusik2000image} or shape/appearance priors~\cite{blanz1999morphable,loper2015smpl,cootes2001active,lensch2003image,zhang1999shape}.
However, the above approaches are either limited to task-specific domains or can only provide weak supervision due to the sparsity of the 2D features.  
In contrast, as the process of producing 2D images from 3D assets, rendering relates each pixel with the 3D parameters by simulating the physical mechanism of image formulation.
Hence, by inverting a renderer, one can obtain {\it dense} pixel-level supervision for {\it general-purpose} 3D reasoning tasks, which cannot be achieved by conventional approaches.  


However, the rendering process is not differentiable in conventional graphics pipelines.
In particular, standard mesh renderer involves a discrete sampling operation, called {\it rasterization}, which prevents the gradient to be flowed into the mesh vertices.
Since the forward rendering function is highly non-linear and complex, to achieve differentiable rendering, recent advances~\cite{loper2014opendr,kato2018neural} only approximate the backward gradient with hand-crafted functions while directly employing a standard graphics renderer in the forward pass.
While promising results have been shown in the task of image-based 3D reconstruction, the inconsistency between the forward and backward propagations may lead to uncontrolled optimization behaviors and limited generalization capability to other 3D reasoning tasks.	
We show in Section~\ref{sec:result_fitting} that such mechanism would cause problematic situations in image-based shape fitting where the 3D parameters cannot be efficiently optimized.

In this paper, instead of studying a better form of rendering gradient, we attack the key problem of differentiating the forward rendering function.
Specifically, we propose a {\it truly differentiable} rendering framework that is able to render a colorized mesh in the forward pass (Figure~\ref{fig:teaser}). 
In addition, our framework can consider a variety of 3D properties, including mesh geometry, vertex attributes (color, normal {\it etc.}), camera parameters and illuminations and is able to flow efficient gradients from pixels to mesh vertices and their attributes.
While being a universal module, our renderer can be plugged into either a neural network or a non-learning optimization framework without parameter tuning. 


The key to our approach is the novel formulation that views rendering as a ``soft" probabilistic process.
Unlike the standard rasterizer, which only selects the color of the closest triangle in the viewing direction (Figure~\ref{fig:teaser} below), we propose that all triangles have probabilistic contributions to each rendered pixel, which can be modeled as probability maps on the screen space.
While conventional rendering pipelines merge shaded fragments in a one-hot manner, we propose a differentiable aggregation function that fuses the per-triangle color maps based on the probability maps and the triangles' relative depths to obtain the final rendering result (Figure~\ref{fig:teaser} upper). 
The novel aggregating mechanism enables our renderer to flow gradients to all mesh triangles, including the occluded ones. 
In addition, our framework can propagate supervision signals from pixels to far-range triangles because of its probabilistic formulation. 
We call our framework {\it Soft Rasterizer (SoftRas)} as it ``softens" the discrete rasterization to enable differentiability.


Thanks to the consistent forward and backward propagations, SoftRas is able to provide high-quality gradient flows that supervise a variety of tasks on image-based 3D reasoning.
To evaluate the performance of SoftRas, we show  applications in 3D unsupervised single-view mesh reconstruction and image-based shape fitting (Figure~\ref{fig:teaser2},  Section~\ref{sec:result_recon} and~\ref{sec:result_fitting}).
In particular, as SoftRas provides strong error signals to the mesh generator simply based on the rendering loss, one can achieve mesh reconstruction from a single image without any 3D supervision. 
To faithfully texture the mesh, we further propose a novel approach that extracts representative colors from input image and formulates the color regression as a classification problem.
Regarding the task of image-based shape fitting, we show that our approach is able to (1) handle occlusions using the aggregating mechanism that considers the probabilistic contributions of all triangles; and (2) provide much smoother energy landscape, compared to other differentiable renderers, that avoids local minima by using the smooth rendering (Figure~\ref{fig:teaser2} left).
Experimental results demonstrate that our approach significantly outperforms the state-of-the-arts both quantitatively and qualitatively.

\section{Related Work}
\label{sec:related_work}

\begin{figure*}[t!]
  	\centering
	\includegraphics[width=1.0\linewidth]{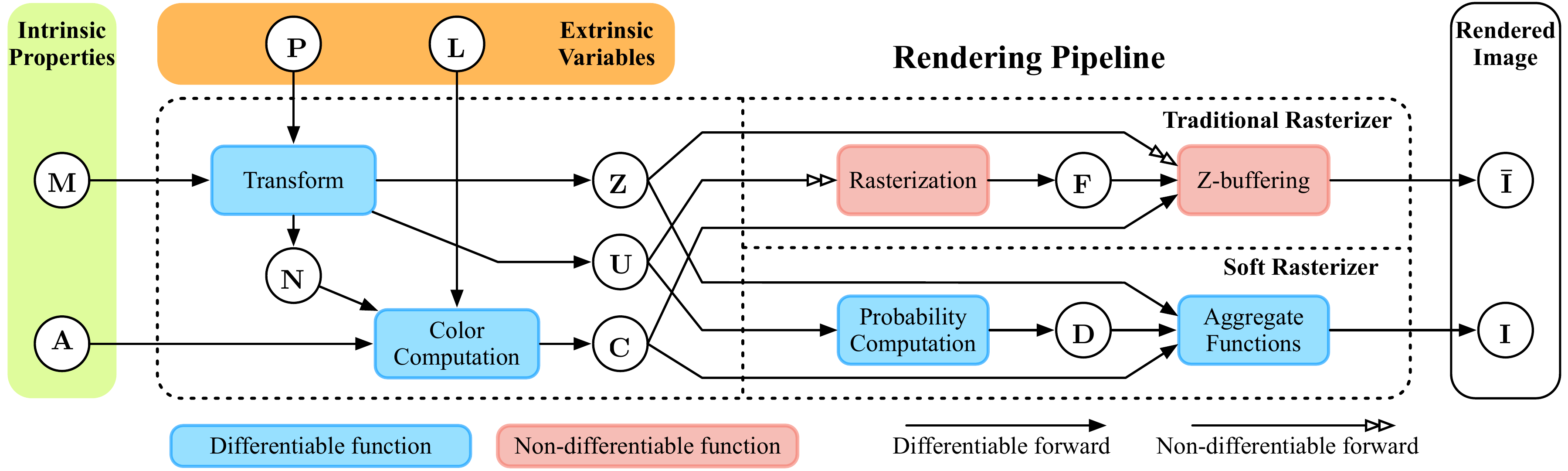}
\caption{Comparisons between the standard rendering pipeline (upper branch) and our rendering framework (lower branch). 	
}
  	\label{fig:computeGraph}
\end{figure*}

\paragraph{Differentiable Rendering.}

To relate the changes in the observed image with that in the 3D shape manipulation, a number of existing techniques have utilized the derivatives of rendering~\cite{gkioulekas2013inverse,gkioulekas2016evaluation,mansinghka2013approximate}.
Recently, Loper and Black~\cite{loper2014opendr} introduce an approximate differentiable renderer which generates derivatives from projected pixels to the 3D parameters.
Kato et al.~\cite{kato2018neural} propose to approximate the backward gradient of rasterization with a hand-crafted function to achieve differentiable rendering.
More recently, Li et al.~\cite{Li:2018:DMC} introduce a differentiable ray tracer to realize the differentiability of secondary rendering effects.
Recent advances in 3D face reconstruction~\cite{richardson2017learning, tewari2017mofa, tewari2018self,tran2018nonlinear,genova2018unsupervised}, material inference \cite{liu2017material, deschaintre2018single} and other 3D reconstruction tasks~\cite{zienkiewicz2016real, rezende2016unsupervised, nalbach2017deep, henderson18bmvc, kundu20183d, nguyen2018rendernet} have leveraged some other forms of differentiable rendering layers to obtain gradient flows in the neural networks.
However, these rendering layers are usually designed for special purpose and thus cannot be generalized to other applications. 
In this paper, we focus on a general-purpose differentiable rendering framework that is able to directly render a given mesh using differentiable functions instead of only approximating the backward derivatives.


\paragraph{Image-based 3D Reasoning.}
2D images are widely used as the media for reasoning 3D properties. 
In particular, image-based reconstruction has received the most attentions.
Conventional approaches mainly leverage the stereo correspondence based on the multi-view geometry \cite{hartley2003multiple,furukawa2010accurate} but is restricted to the coverage provided by the multiple views.
With the availability of large-scale 3D shape dataset \cite{chang2015shapenet}, learning-based approaches~\cite{wang2018pixel2mesh,groueix2018atlasnet,huang2018deep} are able to consider single or few images thanks to the shape prior learned from the data.
To simplify the learning problem, recent works reconstruct 3D shape via predicting intermediate 2.5D representations, such as depth map \cite{liu2016learning}, image collections~\cite{kanazawa2018learning}, displacement map \cite{huynh2018mesoscopic} or normal map \cite{qi2018geonet, wang2015designing}. 
Pose estimation is another key task to understanding the visual environment. 
For 3D rigid pose estimation, while early approaches attempt to cast it as classification problem \cite{tulsiani2015viewpoints}, recent approaches \cite{kendall2015posenet,xiang2018posecnn} can directly regress the 6D pose by using deep neural networks.
Estimating the pose of non-rigid objects, e.g. human face or body, is more challenging. 
By detecting the 2D key points, great progress has been made to estimate the 2D poses~\cite{masi2016pose,cao2018openpose,wei2016convolutional}.
To obtain 3D pose, shape priors~\cite{blanz1999morphable,loper2015smpl} have been incorporated to minimize the shape fitting errors in recent approaches~\cite{bogo2016keep,cao2018openpose,kanazawa2018end,blanz2003face}.
Our proposed differentiable renderer can provide dense rendering supervision to 3D properties, benefitting a variety of image-based 3D reasoning tasks.

\section{Soft Rasterizer}
\label{sec:rasterizer}


\subsection{Differentiable Rendering Pipeline}
\label{sec:pipeline}



As shown in Figure~\ref{fig:computeGraph}, we consider both extrinsic variables (camera $\camera$ and lighting conditions $\lighting$) that define the environmental settings, and intrinsic properties (triangle meshes $\geometry$ and per-vertex appearance $\appearance$, including color, material \textit{etc}.) that describe the model-specific properties. 
Following the standard rendering pipeline, one can obtain the mesh normal $\normal$, image-space coordinate $\uv$ and view-dependent depths $\depth$ by transforming input geometry $\geometry$ based on camera $\camera$.
With specific assumptions of illumination and material models (e.g. Phong model), we can compute color $\vcolor$ given $\{\appearance, \normal, \lighting\}$.
These two modules are \textit{naturally differentiable}.
However, the subsequent operations including the \textit{rasterization} and \textit{z-buffering} in the standard graphics pipeline (Figure~\ref{fig:computeGraph} red blocks) are not differentiable with respect to $\uv$ and $\depth$ due to the discrete sampling operations.
\vspace{-10pt}
\paragraph{Our differentiable formulation.}


We take a different perspective that the rasterization can be viewed as \textit{binary masking} that is determined by the relative positions between the pixels and triangles, while z-buffering merges the rasterization results $\fragment$ in a pixel-wise \textit{one-hot} manner based on the relative depths of triangles. 
The problem is then formulated as modeling the discrete binary masks and the one-hot merging operation in a soft and differentiable manner.
To achieve this, we propose two major components, namely probability maps $\{\Dmap_j\}$ that model the probability of each pixel staying inside a specific triangle $\face_j$ and aggregate function $\aggregate(\cdot)$ that fuses per-triangle color maps based on $\{\Dmap_j\}$ and the relative depths among triangles.

\vspace{-5pt}
\begin{figure}[h]
	\begin{subfigure}[b]{.24\linewidth}
    	\centering
    	\includegraphics[width=\linewidth]{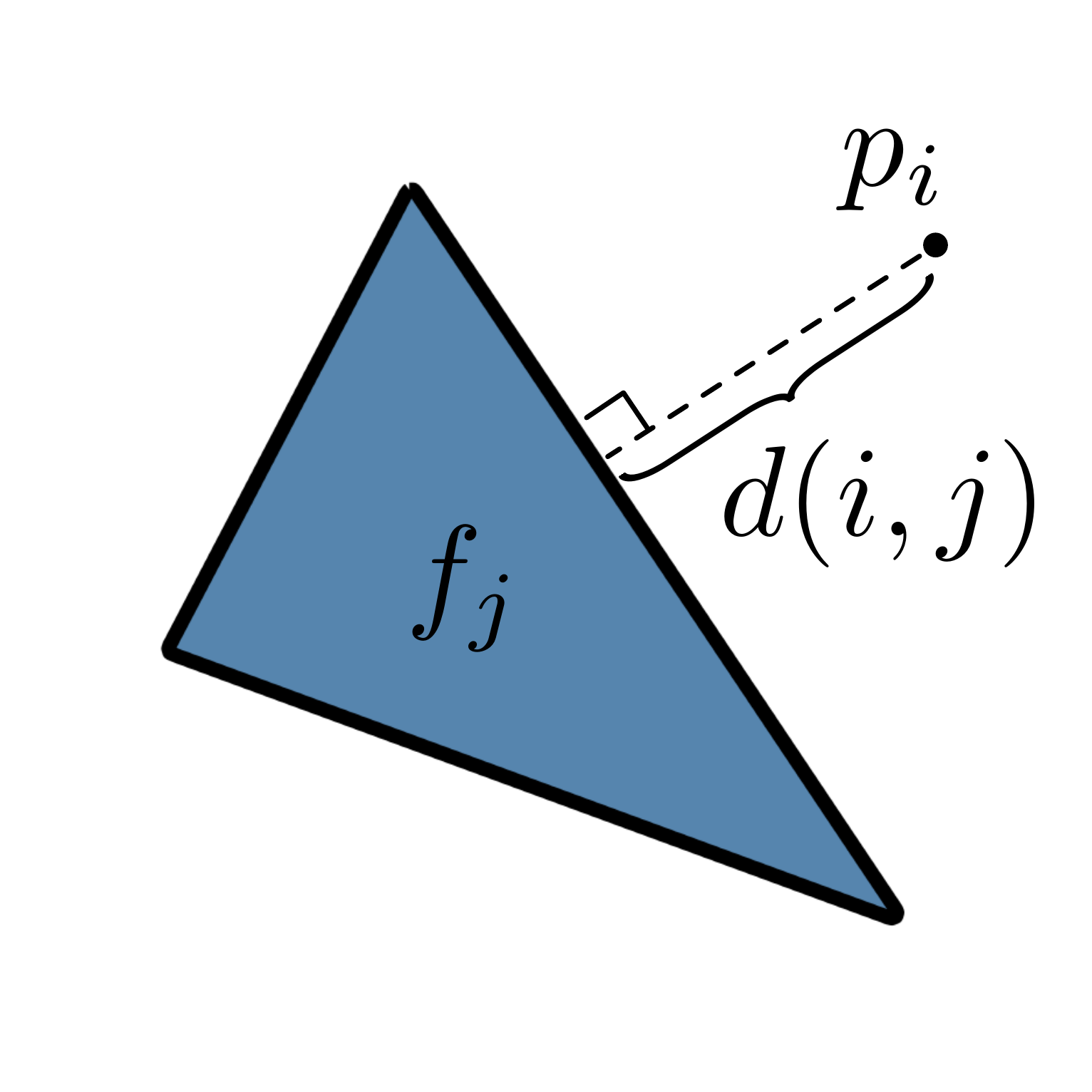}\\
    	\caption{ground truth}
	\end{subfigure}
	\begin{subfigure}[b]{.24\linewidth}
    	\centering
    	\includegraphics[width=\linewidth]{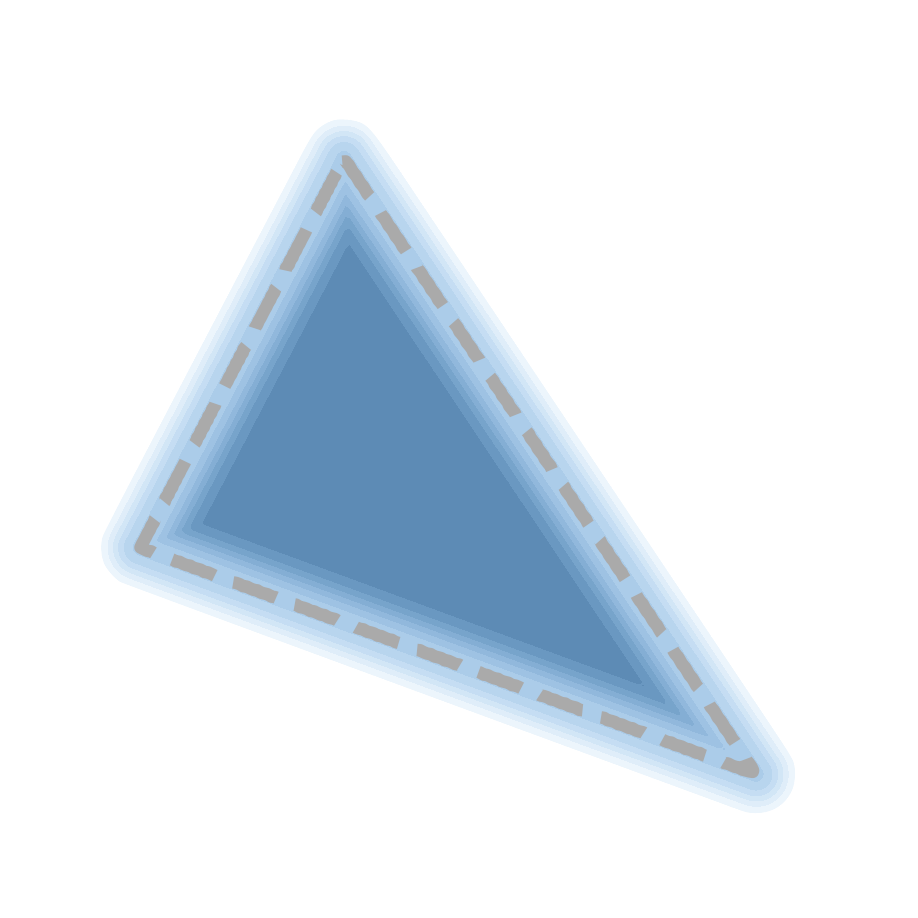}\\
    	\caption{$\sigma = 0.003$}
	\end{subfigure}
	\begin{subfigure}[b]{.24\linewidth}
    	\centering
    	\includegraphics[width=\linewidth]{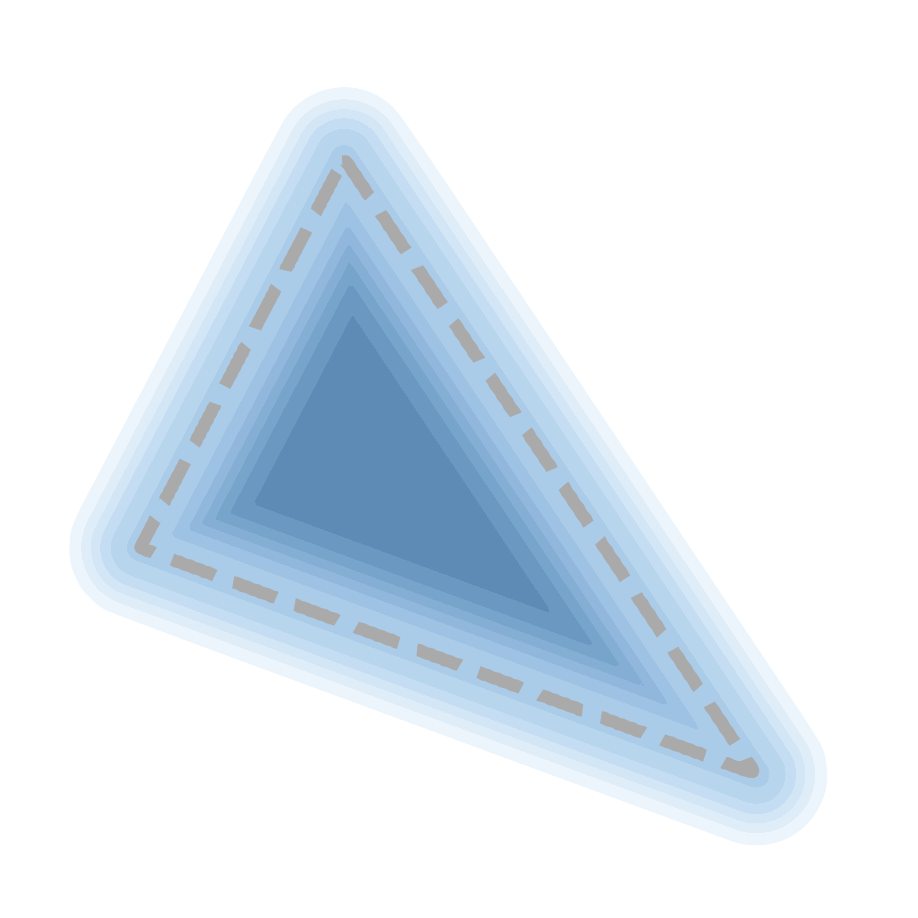}\\
    	\caption{$\sigma = 0.01$}
	\end{subfigure}
	\begin{subfigure}[b]{.24\linewidth}
    	\centering
    	\includegraphics[width=\linewidth]{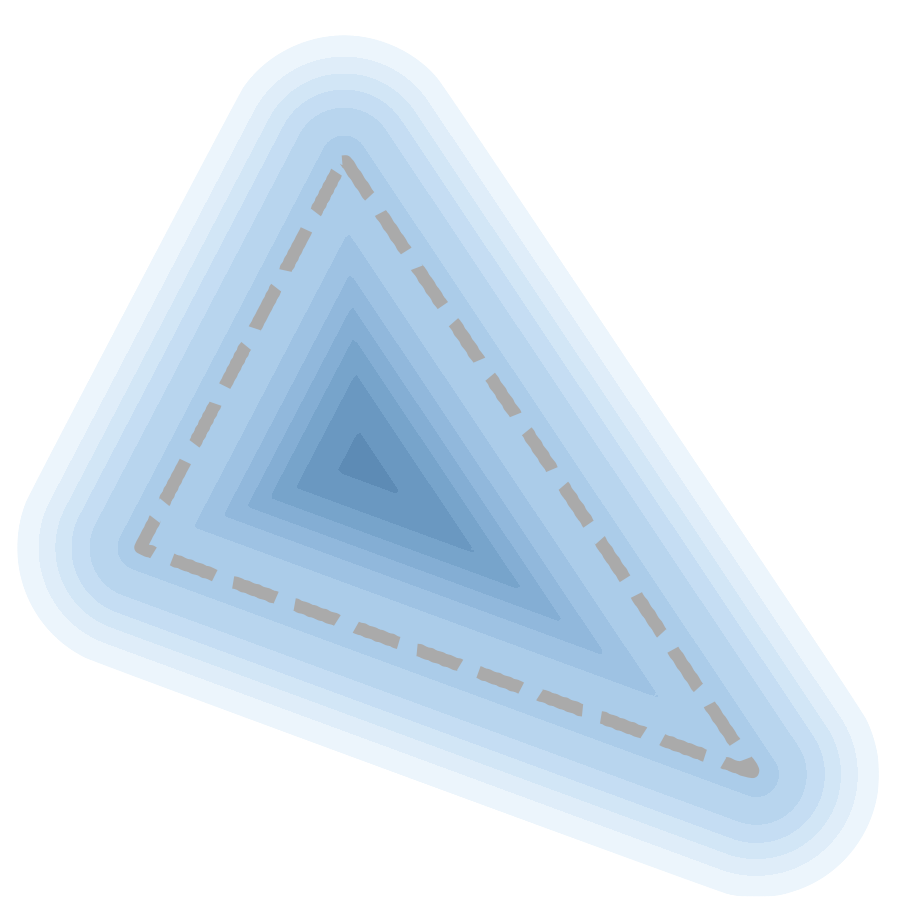}\\
    	\caption{$\sigma = 0.03$}
	\end{subfigure}
	
	\vspace{-5pt}

   	\caption{Probability maps of a triangle under Euclidean metric. (a) definition of pixel-to-triangle distance; (b)-(d) probability maps generated with different $\sigma$.}
   	\label{fig:triangle}

\end{figure}

\subsection{Probability Map Computation}
\label{sec:prob_map}


We model the influence of triangle $\face_j$ on image plane by probability map $\Dmap_j$.
To estimate the probability of $\Dmap_j$ at pixel $\pixel_i$, the function is required to take into account both the relative position and the distance between $\pixel_i$ and $\Dmap_j$. To this end, we define $\Dmap_j$ at pixel $p_i$ as follows:
\vspace{-5pt}
\begin{equation}
\Dmap_j^i = sigmoid(\sig_j^i \cdot \frac{d^2(i,j)}{\sigma}),
\label{eqn:di}
\end{equation}
\vspace{-5pt}

\noindent where $\sigma$ is a positive scalar that controls the sharpness of the probability distribution while $\sig_j^i$ is a sign indicator $\sig_j^i = \{+1, \mathrm{if} \ \pixel_i \in \face_j; -1, \mathrm{otherwise}\}$. 
We set $\sigma$ as $1\times{10}^{-4}$ unless otherwise specified. 
$\dist(i,j)$ is the closest distance from $p_i$ to $\face_j$'s edges. A natural choice for $\dist(i,j)$ is the Euclidean distance. However, other metrics, such as barycentric or $l_1$ distance, can be used in our approach. 


\nothing{
\begin{align*}
	\sig_{ij} =
	\begin{cases}
	+1 & \mathrm{if} \ \pixel_i \in \face_j\\
	-1 & \mathrm{otherwise}.
	\end{cases}
\end{align*}
}

Intuitively, by using the {\it sigmoid} function, Equation~\ref{eqn:di} normalizes the output to $\left(0, 1\right)$, which is a faithful continuous approximation of binary mask with boundary landed on 0.5.
In addition, the sign indicator maps pixels inside and outside $\face_j$ to the range of $\left(0.5, 1\right)$ and $\left(0, 0.5 \right)$ respectively. 
Figure~\ref{fig:triangle} shows $\Dmap_j$ of a triangle with varying $\sigma$ using Euclidean distance.
Smaller $\sigma$ leads to sharper probability distribution while larger $\sigma$ tends to blur the outcome. This design allows controllable influence for triangles on image plane. 
As $\sigma\rightarrow 0$, the resulting probability map converges to the exact shape of the triangle, enabling our probability map computation to be a generalized form of traditional rasterization.


\nothing{
	where $\D_j^i$ is the probability value at the $i$-th pixel $\pixel_i$ of $\Dmap_j$; $\dist(i,j)$ returns the shortest distance from $p_i$ to the edges of $\face_j$ (see Figure~\ref{fig:triangle}(a)); $\sigma$ is a positive hyperparameter that controls the sharpness of the probability distribution while $\sig_{ij}$ is a signed indicator whose response depends on the relative position between $\pixel_i$ and $\face_j$: 
	where $\D_j^i$ is the probability value at the $i$-th pixel $\pixel_i$ (scan line order) of $\Dmap_j$; $\dist(i,j)$ returns the shortest distance from $p_i$ to $\face_i$; $\sigma$ is a positive hyperparameter that controls the sharpness of the probability distribution while $\sig{ij}$ is a signed indicator whose response depends on the relative position between $\pixel_i$ and $\face_j$: \shichen{could be more detailed}
}

\subsection{Aggregate Function}
\label{sec:aggregate}


For each mesh triangle $\face_j$, we define its color map $\equColor_j$ at pixel $\pixel_i$ on the image plane by interpolating vertex color using barycentric coordinates. We clip and normalize the barycentric coordinates to $\left[0,1 \right]$ for $\pixel_i$ outside of $\face_j$.
We then propose to use an aggregate function $\aggregate(\cdot)$ to merge color maps $\{\equColor_j\}$ to obtain rendering output $\equImage$ based on $\{\Dmap_j\}$ and the relative depths $\{\equZ_j\}$.
Inspired by the softmax operator, we define an aggregate function $\aggregate_S$ as follows:
\vspace{-4pt}
\begin{equation}
\equImage^i = \aggregate_S(\{\equColor_j\}) = \sum_j \weight_j^i \equColor_j^i + \weight_b^i \backColor,
\label{equ:aggr}
\vspace{-6pt}
\end{equation}
\noindent where $\backColor$ is the background color; the weights $\{\weight_j\}$ satisfy $\sum_j \weight_j^i + \weight_b^i = 1$ and are defined as:
\vspace{-4pt}
\begin{equation}
\weight_j^i = \frac{\equDis_j^i \exp({\equZ_j^i} / {\equGamma})}{\sum_k \equDis_k^i \exp({\equZ_k^i} / {\equGamma}) + \exp({\epsilon} / {\equGamma})},
\label{equ:aggr_weights}
\end{equation}
\noindent where $\equZ_j^i$ denotes the normalized inverse depth of the 3D point on $\face_i$ whose 2D projection is $\pixel_i$; $\epsilon$ is small constant that enables the background color while $\equGamma$ (set as $1\times{10}^{-4}$ unless otherwise specified) controls the sharpness of the aggregate function. 
Note that $\weight_j$ is a function of two major variables: $\equDis_j$ and $\equZ_j$. 
Specifically, $\weight_j$ assigns higher weight to closer triangles that have larger $\equZ_j$.
As $\equGamma \rightarrow 0$, the color aggregation function only outputs the color of nearest triangle, which exactly matches the behavior of z-buffering. 
In addition, $\weight_j$ is robust to z-axis translations. 
$\equDis_j$ modulates the $\weight_j$ along the $x$, $y$ directions such that the triangles closer to $\pixel_i$ on screen space will receive higher weight.
Equation~\ref{equ:aggr} also works for shading images when the intrinsic vertex colors are set to constant ones.  
We further explore the aggregate function for silhouettes. 
Note that the silhouette of object is independent from its color and depth map.
Hence, we propose a dedicated aggregation function $\aggregate_O$ for the silhouette based on the binary occupancy:
\vspace{-3pt}
\begin{equation}
\silGT^i = \aggregate_O(\{\Dmap_j\}) = 1- \prod_j (1 - \Dmap_j^i).	
\label{equ:silAggr}
\end{equation}
Intuitively, Equation~\ref{equ:silAggr} models silhouette as the probability of having \textit{at least} one triangle cover the pixel $\pixel_i$. 
Note that there might exist other forms of aggregate functions. 
One alternative option may be using a universal aggregate function $\aggregate_N$ that is implemented as a neural network. 
We provide an ablation study on this regard in Section~\ref{sec:ablation}.



\nothing{	
To achieve this goal, we first analyze the working principal of standard rendering pipeline.
The standard rasterizer synthesizes silhouette image by following a strict logical {\it or} operation.
In particular, a pixel that is covere by any triangle on the image plane will be rendered as the interior of the object in the output silhouette image.
We assume that the probability of each triangle face that covers a specific pixel is independent and identically distributed. \weikai{Not sure if i.i.d is guaranteed.}

Figure~\ref{fig:framework} demonstrates the generated silhouette using our \model \ with $\sigma = 3\times10^{-5}$. 
As shown in the result, \modelshort is able to faithfully approximate the ground-truth silhouette without losing fine details.
However, unlike the standard rasterizer, which includes discrete non-differentiable operations, our \modelshort is entirely differentiable thanks to its continuous definition. 
}

\subsection{Comparisons with Prior Works}
\label{sec:method_compare}

\begin{figure}[t!]
  	\centering
	\includegraphics[width=1.0\linewidth]{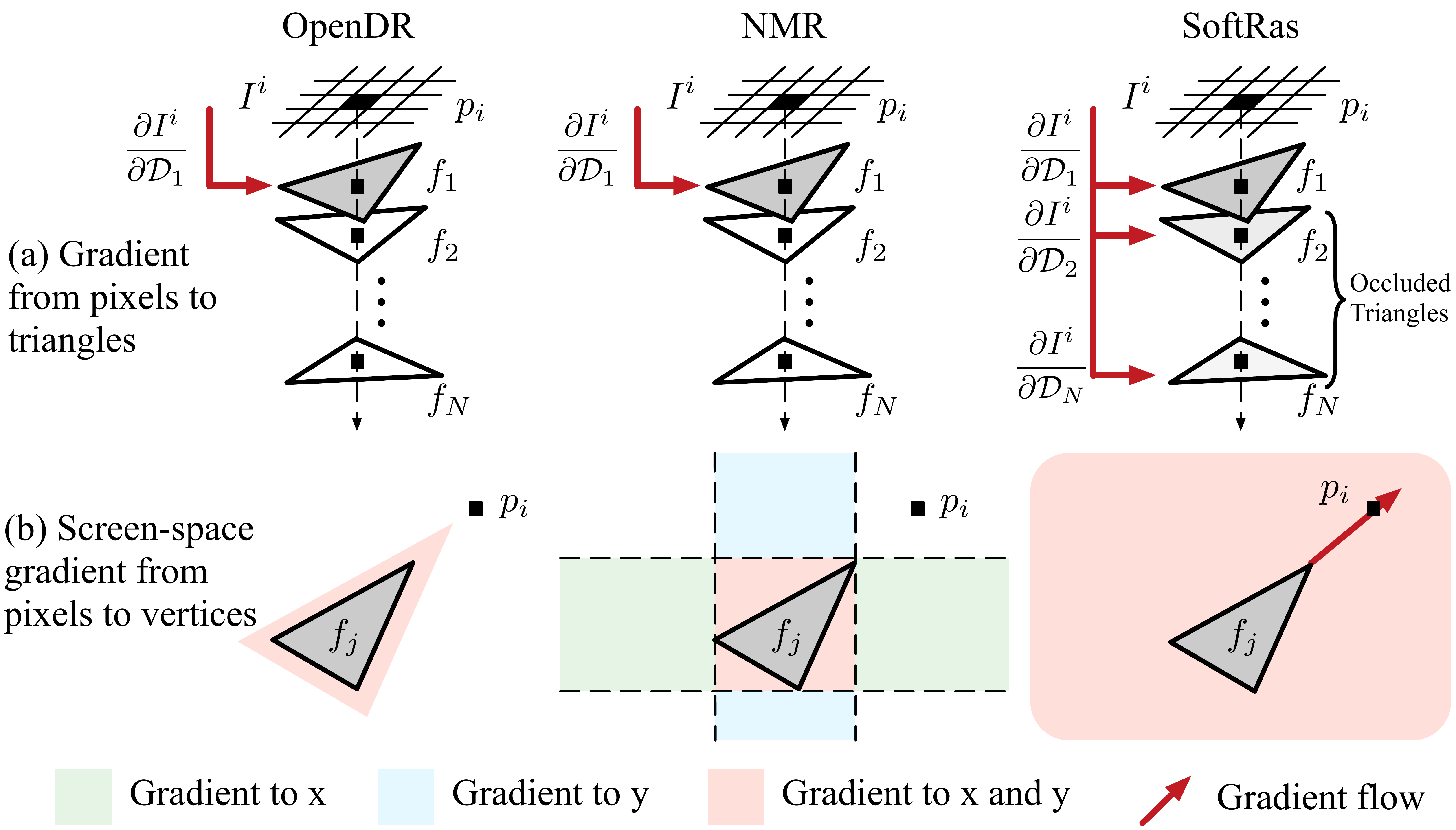}
\caption{Comparisons with prior differentiable renderers in terms of gradient flow. 
}
  	\label{fig:gradient_compare}
\end{figure}


In this section, we compare our approach with the state-of-the-art rasterization-based differential renderers: OpenDR~\cite{loper2014opendr} and NMR~\cite{kato2018neural}, in terms of gradient flows as shown in Figure~\ref{fig:gradient_compare}.
We provide detailed analysis on gradient computation in Appendix~\ref{sec:suppl_gradient}.

\vspace{-3mm}
\paragraph{Gradient from pixels to triangles.}
Since both OpenDR and NMR utilize standard graphics renderer in the forward pass, they have no control over the intermediate rendering process and thus cannot flow gradient into the triangles that are occluded in the final rendered image (Figure~\ref{fig:gradient_compare}(a) left and middle). 
In addition, as their gradients only operate on the image plane, both OpenDR and NMR are not able to optimize the depth value $z$ of the triangles.
In contrast, our approach has full control on the internal variables and is able to flow gradients to invisible triangles and the $z$ coordinates of all triangles through the aggregation function (Figure~\ref{fig:gradient_compare}(a) right).
 

\paragraph{Screen-space gradient from pixels to vertices.}
Thanks to our continuous probabilistic formulation, in our approach, the gradient from pixel $p_j$ in screen space can flow gradient to all distant vertices (Figure~\ref{fig:gradient_compare}(b) right).  
However, for OpenDR, a vertex can only receive gradients from neighboring pixels within a close distance due to the local filtering operation (Figure~\ref{fig:gradient_compare}(b) left).
Regarding NMR, there is no gradient defined from the pixels inside the white regions with respect to the triangle vertices ((Figure~\ref{fig:gradient_compare}(b) middle).
In contrast, our approach does not have such issue thanks to our orientation-invariant formulation.


\section{Image-based 3D Reasoning}

With direct gradient flow from image to 3D properties, our differentiable rendering framework enables a variety of tasks on 3D reasoning. 

\subsection{Single-view Mesh Reconstruction}
\label{sec:overview}

\begin{figure}[t]
	\centering
	\includegraphics[width=1\linewidth]{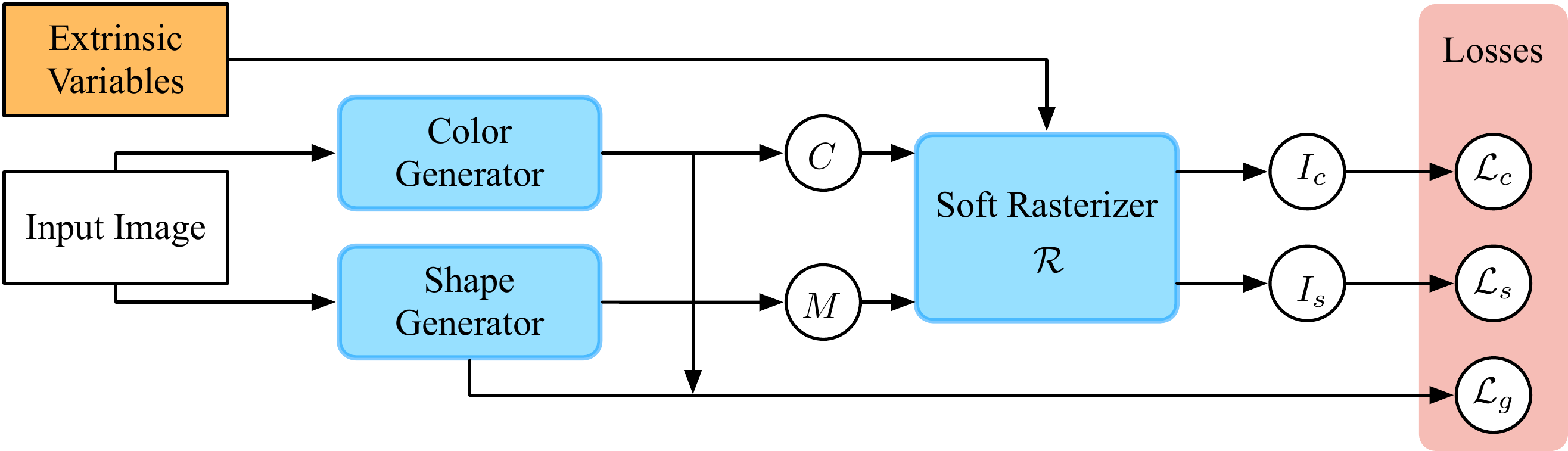}
	\caption{The proposed framework for single-view mesh reconstruction.
		}
	  \label{fig:framework}
	  \vspace{-5pt}
\end{figure}

To demonstrate the effectiveness of \model, we fix the extrinsic variables and evaluate its performance on single-view 3D reconstruction by incorporating it with a mesh generator.
The direct gradient from image pixels to shape and color generators enables us to achieve {\it 3D unsupervised} mesh reconstruction.
Our framework is demonstrated in Figure~\ref{fig:framework}.
Given an input image, our shape and color generators generate a triangle mesh $M$ and its corresponding colors $\equColor$, which are then fed into the soft rasterizer.
The SoftRas layer renders both the silhouette $\equImage_s$ and color image $\equImage_c$ and provide rendering-based error signal by comparing with the ground truths. 
Inspired by the latest advances in mesh learning \cite{kato2018neural,wang2018pixel2mesh}, we leverage a similar idea of synthesizing 3D model by deforming a template mesh.
To validate the performance of \model, the shape generator employ an encoder-decoder architecture  identical to that of \cite{kato2018neural,yan2016perspective}.
The details of the shape and generators are described in Appendix~\ref{sec:suppl_network}.


\begin{figure}[t]
	\centering
	\includegraphics[width=1\linewidth]{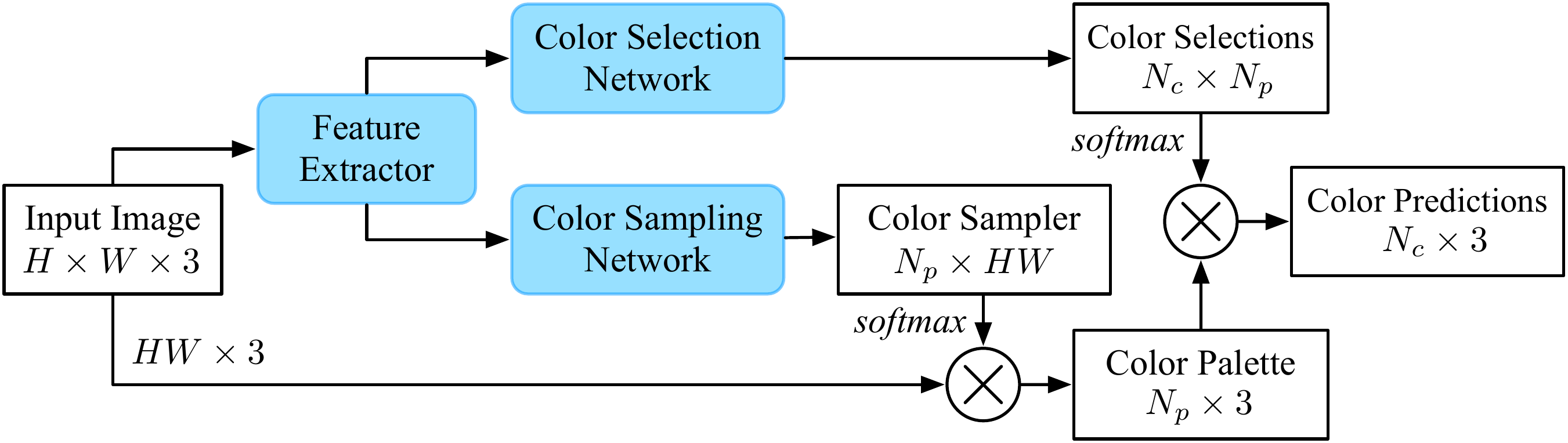}
	\caption{Network structure for color reconstruction. 
		}
	  \label{fig:color_generator}
	  \vspace{-5pt}
\end{figure}


\begin{figure*}[ht!]
    \newlength{\myseqwidth}
    \setlength{\myseqwidth}{.136\linewidth}
    \vspace{-10pt}
	\begin{center}
        \includegraphics[width=1.0\linewidth]{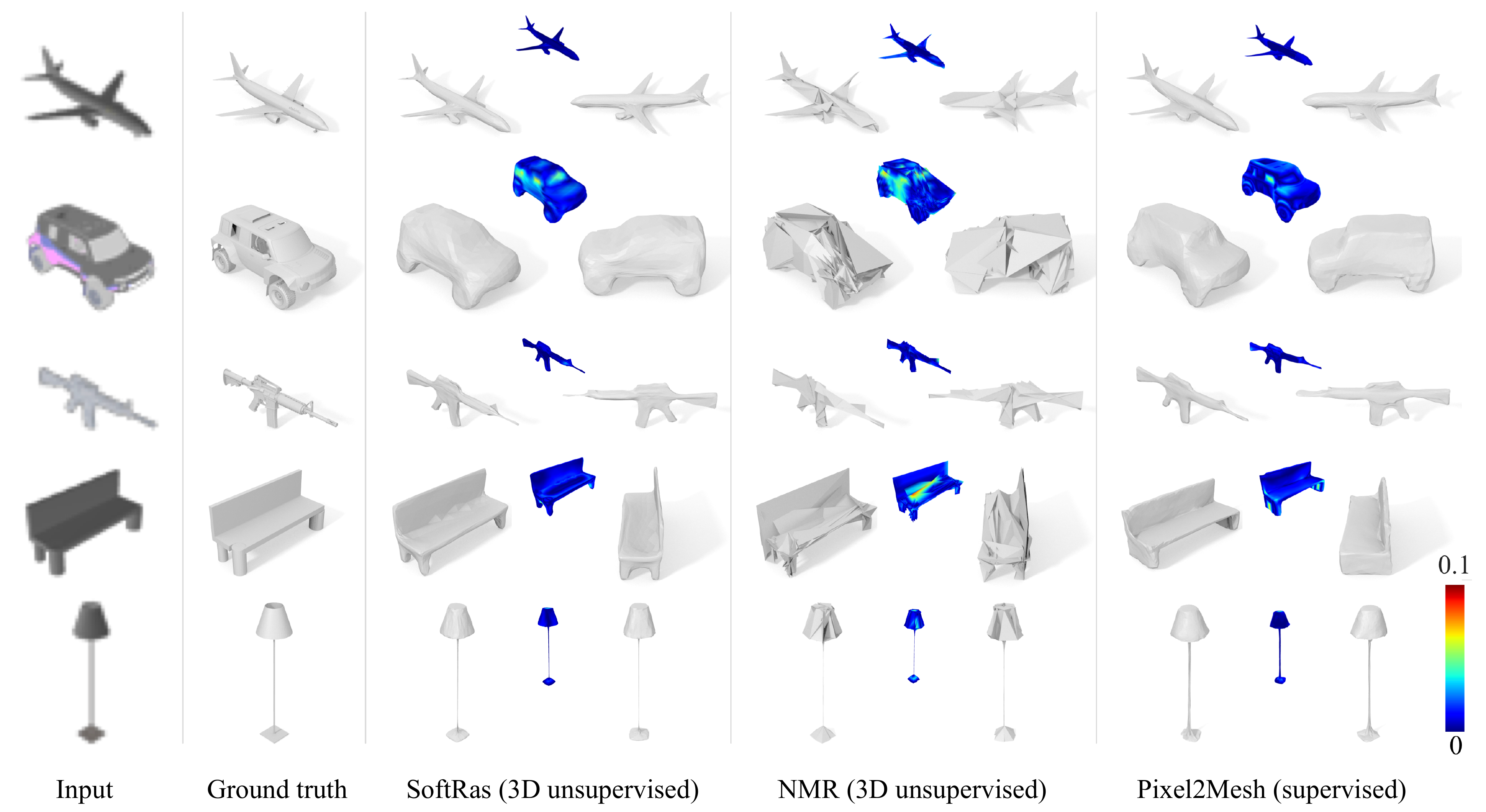}

   \end{center}
   \vspace{-8pt}
   \caption{3D mesh reconstruction from a single image. From left to right, we show input image, ground truth, the results of our method (SoftRas), Neural Mesh Renderer~\cite{kato2018neural} and Pixel2mesh~\cite{wang2018pixel2mesh} -- all visualized from 2 different views. Along with the results, we also visualize mesh-to-scan distances measured from reconstructed mesh to ground truth.}
   \label{fig:vis_compare}
\end{figure*}

\paragraph{Losses.}

The reconstruction networks are supervised by three losses: silhouette loss $\loss_s$, color loss $\loss_c$ and geometry loss $\loss_g$. 
Let $\sil$ and $\silGT$ denote the predicted and the ground-truth silhouette respectively. The silhouette loss is defined as 
$
\loss_s = 1 - \frac{||\sil \otimes \silGT||_1}{||\sil \oplus \silGT - \sil \otimes \silGT||_1},
\label{equ:loss_iou}
$
where $\otimes$ and $\oplus$ are the element-wise product and sum operators respectively. 
The color loss is measured as the $l_1$ norm between the rendered and input image: $\loss_c = || \rgb - \rgbGT ||_1$.
To achieve appealing visual quality, we further impose a geometry loss $\loss_g$ that regularizes the Laplacian of both shape and color predictions.
The final loss is a weighted sum of the three losses:
\vspace{-4pt}
\begin{equation}
\loss = \loss_{s} + \lambda \loss_{c} + \mu \loss_{g}
\label{equ:loss_all}.
\end{equation}


\subsubsection{Color Reconstruction}
\label{sec:color}

Instead of directly regressing the color value, our color generator formulates color reconstruction as a classification problem that learns to reuse the pixel colors in the input image for each sampling point. 
Let $N_c$ denote the number of sampling points on $M$ and $H, W$ be the height and width of the input image respectively. 
However, the computational cost of a naive color selection approach is prohibitive, i.e. $O(HWN_c)$. 
To address this challenge, we propose a novel approach to colorize mesh using a color palette, as shown in Figure~\ref{fig:color_generator}. 
Specifically, after passing input image to a neural network, the extracted features are fed into (1) a sampling network that samples the representative colors for building the palette; and (2) a selection network that combines colors from the palette for texturing the sampling points. 
The color prediction is obtained by multiplying the color selections with the learned color palette.
Our approach reduces the computation complexity to  $O(N_d(HW + N_c))$, where $N_p$ is the size of color palette.
With a proper setting of $N_p$, one can significantly reduce the computational cost while achieving sharp and accurate color recovery.

\subsection{Image-based Shape Fitting}
\label{sec:fitting}

Image-based shape fitting has a fundamental impact in various tasks, such as pose estimation, shape alignment, model-based reconstruction, {\it etc}.
Yet without direct correlation between image and 3D parameters, conventional approaches have to rely on coarse correspondences, e.g. 2D joints \cite{bogo2016keep} or feature points \cite{pavlakos2017coarse}, to obtain supervision signals for optimization.
In contrast, SoftRas can directly back-propagate pixel-level errors to 3D properties, enabling dense image-to-3D correspondence for high-quality shape fitting.
However, a differentiable renderer has to resolve two challenges in order to be readily applicable.
(1) {\it occlusion awareness}: the occluded portion of 3D model should be able to receive gradients in order to handle large pose changes.  
(2) {\it far-range impact}: the loss at a pixel should have influence on distant mesh vertices, which is critical to dealing with local minima during optimization.
While prior differentiable renderers \cite{kato2018neural,loper2014opendr} fail to satisfy these two criteria, our approach handles these challenges simultaneously.
(1) Our aggregate function fuses the probability maps from all triangles, enabling the gradients to be flowed to all vertices including the occluded ones.
(2) Our soft approximation based on probability distribution allows the gradient to be propagated to the far end while the size of receptive field can be well controlled (Figure~\ref{fig:triangle}).
To this end, our approach can faithfully solve the image-based shape fitting problem by minimizing the following energy objective:

\vspace{-2mm}
\begin{equation}
	\argmin_{\rho,\theta,t}||R(M(\rho,\theta,t)) - \equImage_{t}||_2,
\end{equation} 

\noindent where $R(\cdot)$ is the rendering function that generates a rendered image $\equImage$ from mesh $M$, which is parametrized by its pose $\theta$, translation $t$ and non-rigid deformation parameters $\rho$.
The difference between $\equImage$ and the target image $\equImage_t$ provides strong supervision to solve the unknowns $\{\rho, \theta, t\}$.



\section{Experiments}
\label{sec:result}

In this section, we perform extensive evaluations on our framework. 
We also include more visual evaluations in the appendix.

\begin{table*}[h!]
\resizebox{\textwidth}{!}{
\centering
\begin{tabular}{ccccccccccccccc}
\hline
Category    & Airplane & Bench  & Dresser & Car    & Chair  & Display & Lamp   & Speaker & Rifle & Sofa & Table & Phone & Vessel & Mean       \\ 
\hline
retrieval~\cite{yan2016perspective}   & 0.5564   & 0.4875 & 0.5713  & 0.6519 & 0.3512 & 0.3958  & 0.2905 & 0.4600   & 0.5133 & 0.5314  & 0.3097 & 0.6696 & 0.4078  & 0.4766       \\
voxel~\cite{yan2016perspective} & 0.5556   & 0.4924 & 0.6823  & 0.7123 & 0.4494 & 0.5395  & 0.4223 & 0.5868   & 0.5987 & 0.6221  & 0.4938 & 0.7504 & 0.5507  & 0.5736       \\
NMR~\cite{kato2018neural}        & 0.6172   & 0.4998 & 0.7143  & 0.7095 & 0.4990 & 0.5831  & 0.4126 & 0.6536   & 0.6322 & 0.6735  & 0.4829 & 0.7777 & 0.5645  & 0.6015       \\
\hline
Ours (sil.)     & 0.6419   & 0.5080 & 0.7116  & 0.7697 & 0.5270 & 0.6156  & \bf{0.4628} & 0.6654   & \bf{0.6811} & 0.6878  & 0.4487 & 0.7895 & 0.5953  &  0.6234       \\ 
Ours (full)     & \bf{0.6670}   & \bf{0.5429} & \bf{0.7382}  & \bf{0.7876} & \bf{0.5470} & \bf{0.6298}  & 0.4580 &   \bf{0.6807}   & 0.6702 & \bf{0.7220}  & \bf{0.5325} & \bf{0.8127} & \bf{0.6145} & \bf{0.6464}     \\ 
\hline
\end{tabular}}

\vspace{-5pt}
\caption{Comparison of mean IoU with other 3D unsupervised reconstruction methods on 13 categories of ShapeNet datasets. 
}
\label{tab:shapenet}
\vspace{-5mm}
\end{table*}

\subsection{Single-view Mesh Reconstruction}
\label{sec:result_recon}

\subsubsection{Experimental Setup}
\label{sec:setup}

\paragraph{Datasets and Evaluation Metrics.} 
We use the dataset provided by \cite{kato2018neural}, which contains 13 categories of objects from ShapeNet~\cite{chang2015shapenet}.
Each object is rendered in 24 different views with image resolution of 64 $\times$ 64. 
For fair comparison, we employ the same train/validate/test split on the same dataset as in \cite{kato2018neural, yan2016perspective}. 
For quantitative evaluation, we adopt the standard reconstruction metric, 3D intersection over union (IoU), to compare with baseline methods.  


\paragraph{Implementation Details.} 
We use the same structure as \cite{kato2018neural, yan2016perspective} for mesh generation. 
\nothing{
\weikai{move the details to supplemental materials.}
The encoder netowrk consists of 3 convolutional layers with kernel size of 5 $\times$ 5 and channels of 64, 128, 256, followed by 3 fully connected layers with hidden layer size of 1024 and 1024.
In particular, we apply batch normalization \cite{ioffe2015batch} and ReLU activation \cite{krizhevsky2012imagenet} after each convolutional layer.
The decoder transforms a 512-dimensional latent code into a displacement vector of length 1926 (coordinates of 642 vertices) by using 3 fully connected layers with hidden size of 1024 and 2048. 
}
Our network is optimized using Adam~\cite{kingma2014adam} with  $\alpha = 1\times{10}^{-4}$, $\beta_1 = 0.9$ and $\beta_2 = 0.999$. 
Specifically, we set $\lambda = 1$ and $\mu = 1\times{10}^{-3}$ 
across all experiments unless otherwise specified. 
We train the network with multi-view images of batch size 64 and implement it using PyTorch. 


\begin{figure}[t]
	\includegraphics[width=\linewidth]{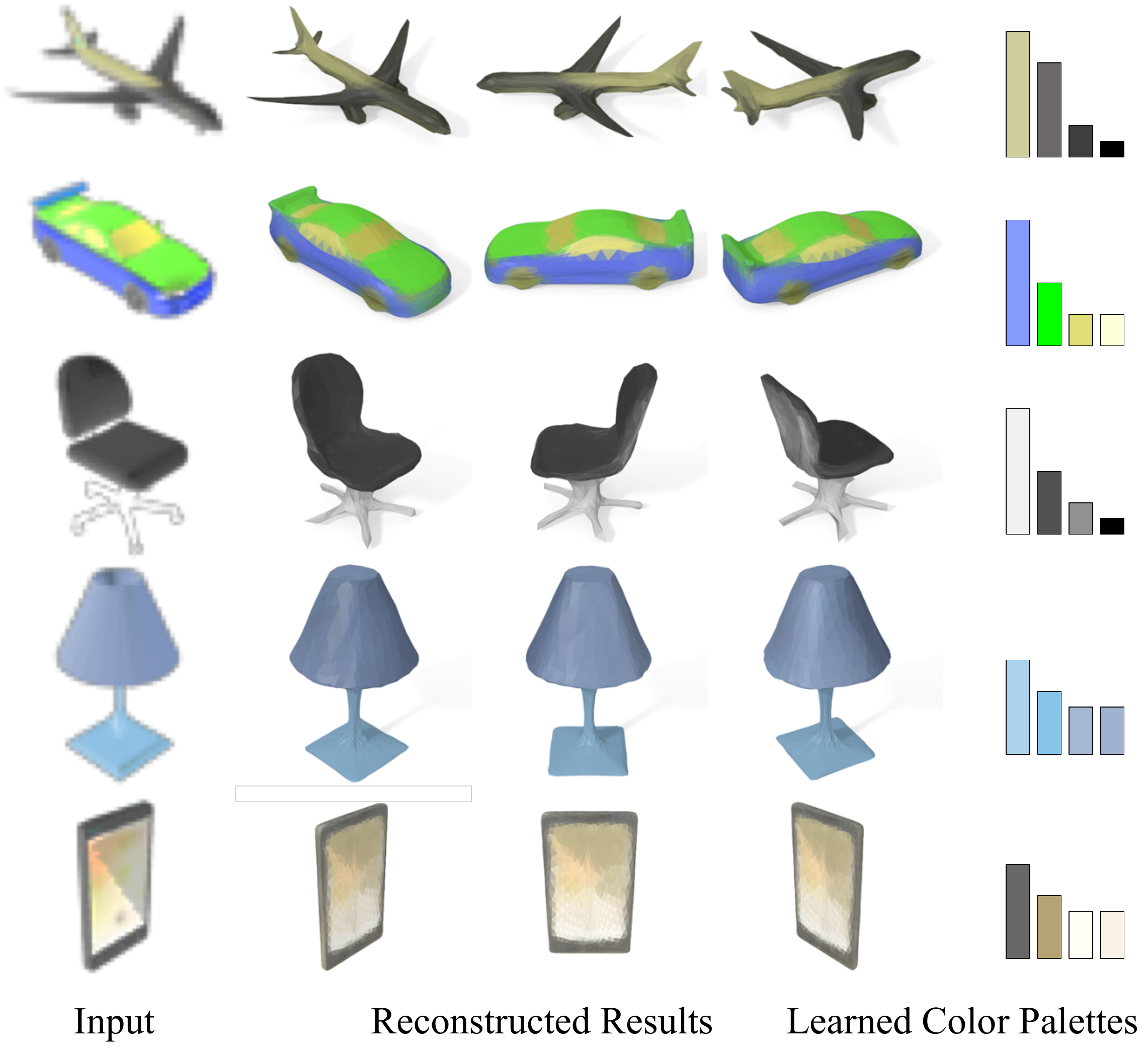}\\
	\vspace{-15pt}
	\caption{Results of colorized mesh reconstruction. The learned principal colors and their usage histogram are visualize on the right. 
}
	\label{fig:color}
	\vspace{-3mm}
\end{figure}

\subsubsection{Qualitative Results}
\label{sec:quali_results}

\nothing{
Qualitative results that we would like to show:
\begin{itemize}
	\item Pure geometry reconstruction. Comparisons with 
		\begin{itemize}
			\item neural 3D mesh renderer
			\item PTN
			\item Multiview ray consistency 
			\item Pixel2Mesh
		\end{itemize}
	\item Color reconstruction. Comparisons with 
		\begin{itemize}
			\item neural 3D mesh renderer
			\item our naive approach of learning color
		\end{itemize}
	\item Reconstruction using real image (train with more views)
	\item Reconstruction in presence of occlusion (optional)
\end{itemize}	
}


\paragraph{Single-view Mesh Reconstruction.} 

We compare the qualitative results of our approach with that of the state-of-the-art supervised~\cite{wang2018pixel2mesh} and 3D unsupervised~\cite{kato2018neural} mesh reconstruction approaches in Figure~\ref{fig:vis_compare}.
Though NMR~\cite{kato2018neural} is able to recover the rough shape, the mesh surface is discontinuous and suffers from a considerable amount of self intersections.
In contrast, our method can faithfully reconstruct fine details of the object, such as the tail of the airplane and the barrel of the rifle, while ensuring smoothness of the surface.
Though trained without 3D supervision, our approach achieves results on par with the supervised method Pixel2Mesh~\cite{wang2018pixel2mesh}. 
In some cases, our approach can generate even more appealing details than that of \cite{wang2018pixel2mesh}, e.g. the bench legs, the airplane engine and the side of the car. 
Mesh-to-scan distance visualization also shows our results achieve much higher accuracy than \cite{kato2018neural} and comparable accuracy with that of \cite{wang2018pixel2mesh}.

\paragraph{Color Reconstruction.}

Our method is able to faithfully recover the mesh color based on the input image.
Figure~\ref{fig:color} presents the colorized reconstruction from a single image and the learned color palettes.
Though the resolution of the input image is rather low ($64 \times 64$), our approach is still able to achieve sharp color recovery and accurately restore the fine details, e.g. the subtle color transition on the body of airplane and the shadow on the phone screen.


\subsubsection{Quantitative Evaluations}
\label{sec:quant_results}

\nothing{
\weikai{TODO Comparisons:
\begin{itemize}
	\item Neural 3D mesh renderer
	\item Perspective transformation network
	\item PTN - retrieval based method
	\item Multiview ray consistency 
	\item Pixel2Mesh (optional)
\end{itemize}	
}
}

We show the comparisons on 3D IoU score with the state-of-the-art approaches in Table~\ref{tab:shapenet}.
We test our approach under two settings: one trained with silhouette loss only (sil.) and the other with both silhouette and shading supervisions (full).
Our approach has significantly outperformed all the other unsupervised methods on all categories.
In addition, the mean score of our best setting has surpassed the state-of-the-art NMR~\cite{kato2018neural} by more than 4.5 points.
As we use the identical mesh generator and same training settings with \cite{kato2018neural}, it indicates that it is the proposed SoftRas renderer that leads to the superior performance. 

\nothing{
T
}

\begin{table}[h!]
\addtolength{\tabcolsep}{1pt}
\centering

\begin{tabular}{C{1.4cm}|C{1.4cm}|C{1.3cm}|c|c}
\hline
\multicolumn{3}{c|}{SoftRas settings}        & \multirow{2}{*}{$\loss_{lap}$} & \multirow{2}{*}{mIoU} \\ \cline{1-3}
distance func. & aggregate func. ($\alpha$) & aggregate func. (color) &                            &                      \\ \hline
Barycentric &   $\aggregate_O$   &    -     &            &     60.8    \\
Euclidean &   $\aggregate_O$    &    -    &       &     62.0    \\
Euclidean &   $\aggregate_O$    &    -   &    $\checkmark$    &    62.4   \\
Euclidean &   $\aggregate_N$    &    -   &    $\checkmark$    &    63.2   \\
Euclidean &   $\aggregate_O$    &   $\aggregate_S$   &    $\checkmark$    &    \textbf{64.6}   \\

\hline
\end{tabular}

\vspace{2pt}
\caption{Ablation study of the regularizer and various forms of distance and aggregate functions. $\aggregate_N$ stands for the aggregation function implemented as a neural network. $\aggregate_S$ and $\aggregate_O$ refer to the aggregation functions defined in Equation~\ref{equ:aggr} and \ref{equ:silAggr} respectively.
}
\label{tab:ablation}
\vspace{-5mm}
\end{table}
\subsubsection{Ablation Study}
\label{sec:ablation}

In this section, we conduct controlled experiments to validate the importance of different components.

\paragraph{Loss Terms and Alternative Functions.} 
In Table~\ref{tab:ablation}, we investigate the impact of Laplacian regularizer and various forms of the distance function (Section~\ref{sec:prob_map}) and the aggregate function.
As the RGB color channel and the $\alpha$ channel (silhouette) have different candidate aggregate functions, we separate their lists in Table~\ref{tab:ablation}.
First, by adding Laplacian constraint, our performance is increased by 0.4 point (62.4 \textit{v.s.} 62.0). In contrast, NMR~\cite{kato2018neural} has reported a negative effect of geometry regularizer on its quantitative results.
The performance drop may be due to the fact that the ad-hoc gradient is not compatible with the regularizer. 
It is optional to have color supervision on the mesh generation. However, we show that adding a color loss can significantly improve the performance (64.6 \textit{v.s.} 62.4) as more information is leveraged for reducing the ambiguity of using silhouette loss only.
In addition, we also show that Euclidean metric usually outperforms the barycentric distance while the aggregate function based on neural network $\aggregate_N$ performs slightly better than the non-parametric counterpart $\aggregate_O$ at the cost of more computations.

\nothing{
We first investigate the influence of the two losses, $\loss_{lap}$ and $\loss_{fl}$, that regularize the geometry property of the generated mesh models. 
Figure~\ref{fig:vis_ablation} shows the results of selectively dropping one of the losses.
As seen from the comparisons, removing the Laplacian loss $\loss_{lap}$ would lead to less smooth surface, e.g. the head of airplane, or even discontinuities, e.g. the hole in the bench seat.
Dropping off the flattening loss $\loss_{fl}$ would severely impairs the smoothness of surface and causes intersecting geometry, e.g. the bench back and the tail part of airplane.
However, when the losses are both present, our full model does not suffer from any of the problems, indicating both of the components are necessary for our final performance.  
}

\nothing{
\paragraph{Training with More Views.} \shichen{remove this part}
As discussed in Section~\ref{sec:quant_results}, the existing dataset only contains limited biased views, resulting in a large degree of ambiguity in the reconstructed models. 
To evaluate the full capability of our model, we render the existing models with more comprehensive views. 
In particular, the new training images are rendered from 120 viewpoints, sampled from 5 elevation and 24 azimuth angles.
As shown in Figure~\ref{fig:vis_120}, by training with more views, our model is capable to generate results much closer to the ground truth.
In particular, the embossment of the table model has been completely removed in the new results, indicating the effectiveness of the silhouette supervision obtained from our \modelshort layer.
}



\nothing{
\paragraph{Reconstruction from Real Image.}
We further evaluate our performance on real images.
To adapt our approach to more viewpoints, we render the existing models using 120 views, sampled from 5 elevation and 24 azimuth angles.
As demonstrated in Figure~\ref{fig:vis_real}, though only trained on synthetic data, our model generalizes well to real images in novel views with faithful reconstruction of fine details, e.g. the tail fins of the fighter aircraft.

}

\subsection{Image-based Shape Fitting}
\label{sec:result_fitting}

\begin{figure}[h]

	\includegraphics[width=1\linewidth]{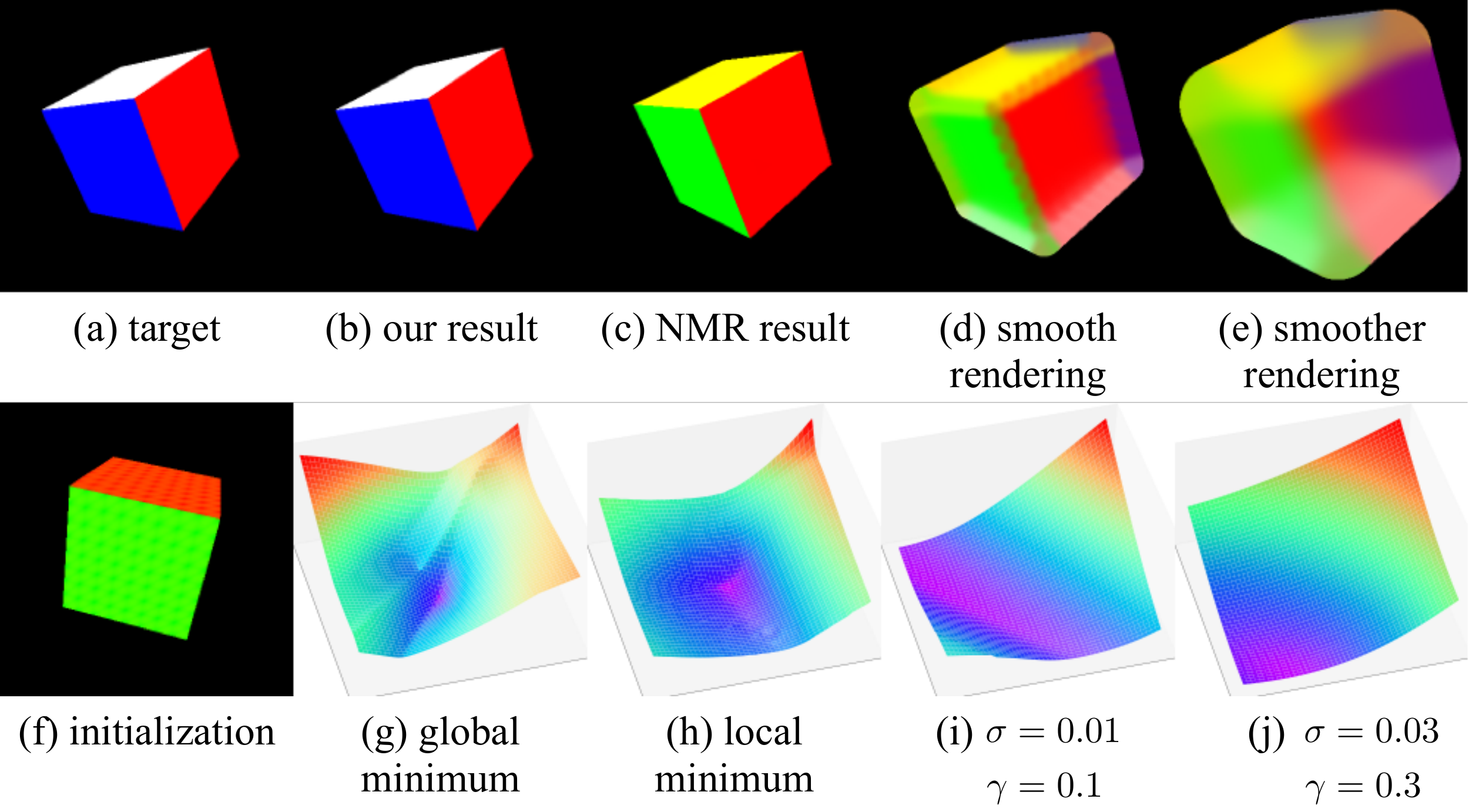}
	\caption{Visualization of loss function landscapes of NMR and SoftRas for pose optimization given target image (a) and initialization (f). SoftRas achieves global minimum (b) with loss landscape (g). NMR is stuck in local minimum (c) with loss landscape (h). At this local minimum, SoftRas produces the smooth and partially transparent rendering (d)(e), which smoothens the loss landscape (i)(j) with larger $\equSigma$ and $\equGamma$, and consequently leads to better minimum.}
	\label{fig:cube}
	\vspace{-3mm}
\end{figure}

\begin{table}[]
\centering
\addtolength{\tabcolsep}{8pt}
\begin{tabular}{l|cc}
\hline
Method   & w/o scheduling & w/ scheduling \\ \hline
baseline & 126.48$\degree$\footnotemark  &  126.48$\degree$   \\
NMR      & 93.40$\degree$  &  80.94$\degree$  \\ \hline
SoftRas  & \textbf{82.80$\degree$}  & \textbf{63.57$\degree$}      \\ \hline
\end{tabular}
\caption{Comparison of cube rotation estimation error with NMR, measured in mean relative angular error.}
\label{tab:cube}
\end{table}

\begin{figure}[h]
	\centering
	\includegraphics[width=\linewidth]{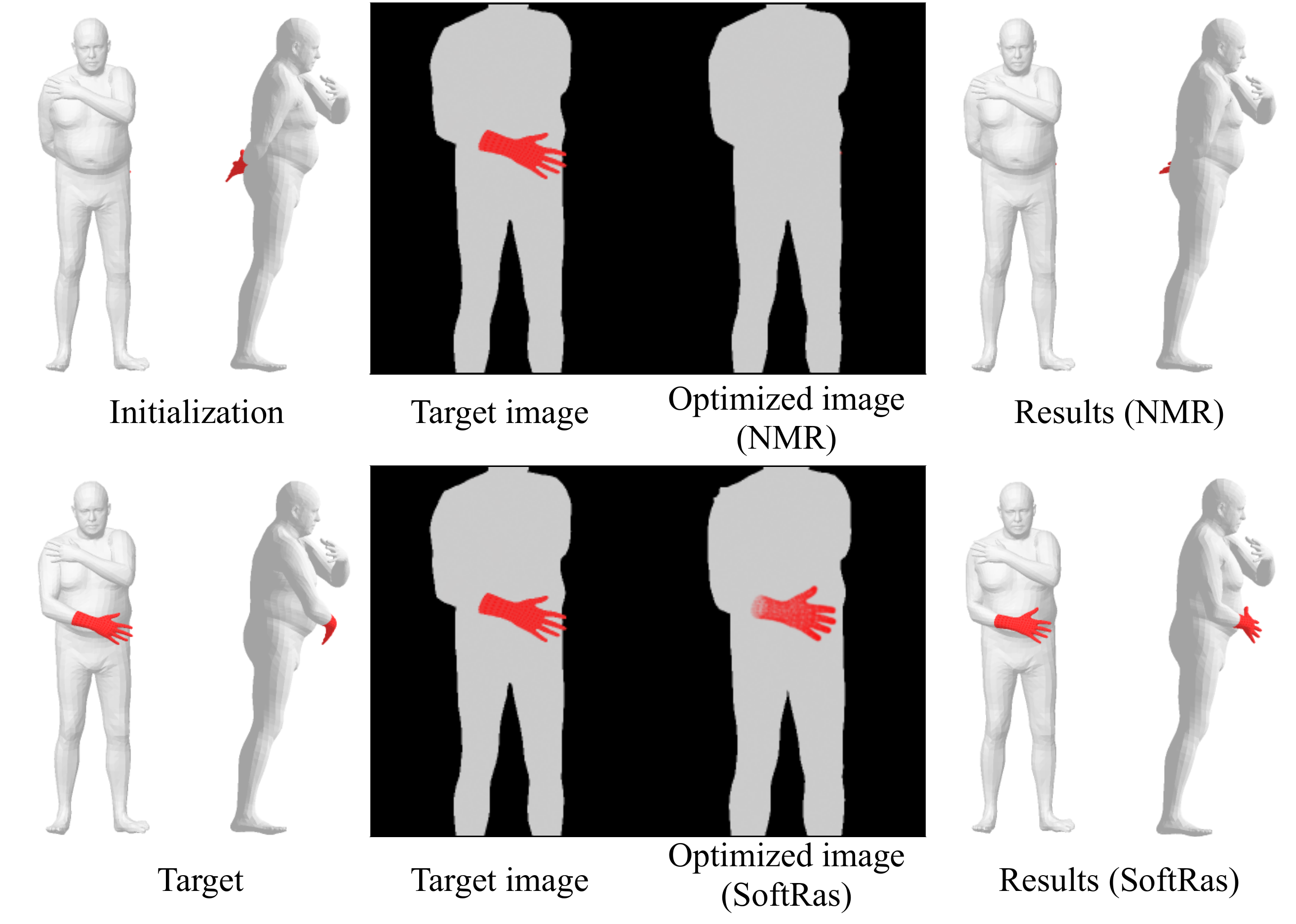}
	\caption{Results for optimizing human pose given single image target. 
		}
	\label{fig:pose_smpl}
\end{figure}


\paragraph{Rigid Pose Fitting.} 
We compare our approach with NMR in the task of rigid pose fitting. In particular, given a colorized cube and a target image, the pose of the cube needs to be optimized so that its rendered result matches the target image.
Despite the simple geometry, the discontinuity of face colors, the non-linearity of rotation and the large occlusions make it particularly difficult to optimize.
As shown in Figure~\ref{fig:cube}, NMR is stuck in a local minimum while our approach succeeds to obtain the correct pose.
\footnotetext{The expectation of uniform-sampled  SO3 rotation angle is $\pi/2+2/\pi$}
The key is that our method produces smooth and partially transparent renderings which ``soften" the loss landscape.
Such smoothness can be controlled by $\equSigma$ and $\equGamma$, which allows us to avoid the local minimum. 
Further, we evaluate the rotation estimation accuracy on synthetic data given 100 randomly sampled initializations and targets. We compare methods w/ and w/o scheduling schemes, and summarize mean relative angle error in Table~\ref{tab:cube}. Without optimization scheduling, our method outperforms the baseline (random estimation) and NMR by 43.68$\degree$ and 10.60$\degree$ respectively, demonstrating the effectiveness of the gradient flows provided by  our method. 
Scheduling is a commonly used technique for solving non-linear optimization problems.
For NMR, we solve with multi-resolution images in 5 levels; while for our method, we set schedule to decay $\equSigma$ and $\equGamma$ in 5 steps.
While scheduling improves both methods, our approach still achieves better accuracy than NMR by 17.37$\degree$, indicating our consistent superiority regardless of using the scheduling strategy.


\paragraph{Non-rigid Shape Fitting.} 
In Figure~\ref{fig:pose_smpl}, we show that SoftRas can provide stronger supervision for non-rigid shape fitting even in the presence of part occlusions. 
We optimize the human body parametrized by SMPL model \cite{loper2015smpl}. 
As the right hand (textured as red) is completely occluded in the initial view, it is extremely challenging to fit the body pose to the target image. 
To obtain correct parameters, the optimization should be able to (1) consider the impact of the occluded part on the rendered image and (2) back-propagate the error signals to the occluded vertices.
NMR~\cite{kato2018neural} fails to move the hand to the right position due to its incapability to handle occlusions.
In comparison, our approach can faithfully complete the task as our novel probabilistic formulation and aggregating mechanism can take all triangles into account while being able to optimize the $z$ coordinates (depth) of the mesh vertices.



\section{Conclusions}
\label{sec:conclusion}

\nothing{
\begin{wrapfigure}{r}{0.25\textwidth}
    \vspace{-2mm}
    \begin{center}
        \includegraphics[width=0.25\textwidth]{limitPh}
    \end{center}
    \vspace{-5mm}
    \vspace{-1mm}
    \label{fig:fail}
\end{wrapfigure}
}

In this paper, we have presented a truly differentiable rendering framework (SoftRas) that is able to directly render a given mesh in a fully differentiable manner.
SoftRas can consider both extrinsic and intrinsic variables in a unified rendering framework and generate efficient gradients flowing from pixels to mesh vertices and their attributes (color, normal, \textit{etc.}).
We achieve this goal by re-formulating the conventional discrete operations including rasterization and z-buffering as differentiable probabilistic processes.
Such novel formulation enables our renderer to provide  more efficient supervision signals, flow gradients to unseen vertices and optimize the $z$ coordinates of mesh triangles, leading to the significant improvements in the tasks of single-view mesh reconstruction and image-based shape fitting.
As a general framework, it would be an interesting future avenue to investigate other possibilities of distance and aggregate functions that might lead to even superior performance.

\nothing{For objects that are not genus-0, e.g. the bench in the second row of Figure~\ref{fig:vis_compare}, Pixel2Mesh generates better results than ours (see the armrest). 
The reason is that as our method can only generate genus-0 surface, synthesizing the armrest will lead to large 2D IoU loss.
In contrast, Pixel2Mesh employs 3D Chamfer distance as loss metric which could strongly penalize the missing of armrest in the reconstructed model.
}

\section*{Acknowledgements}
Hao Li is affiliated with the University of Southern California, the USC Institute for Creative Technologies, and Pinscreen. This research was conducted at USC and was funded by in part by the ONR YIP grant N00014-17-S-FO14, the CONIX Research Center, one of six centers in JUMP, a Semiconductor Research Corporation (SRC) program sponsored by DARPA, the Andrew and Erna Viterbi Early Career Chair, the U.S. Army Research Laboratory (ARL) under contract number W911NF-14-D-0005, Adobe, and Sony. This project was not funded by Pinscreen, nor has it been conducted at Pinscreen or by anyone else affiliated with Pinscreen. The content of the information does not necessarily reflect the position or the policy of the Government, and no official endorsement should be inferred.

{\small
	\bibliographystyle{ieee}
	\bibliography{paper}
}

\ifthenelse{\equal{\forArxiv}{1}}{
	\clearpage
	\appendix
	\begin{figure*}[h]
    \centering
    \includegraphics[width=\textwidth]{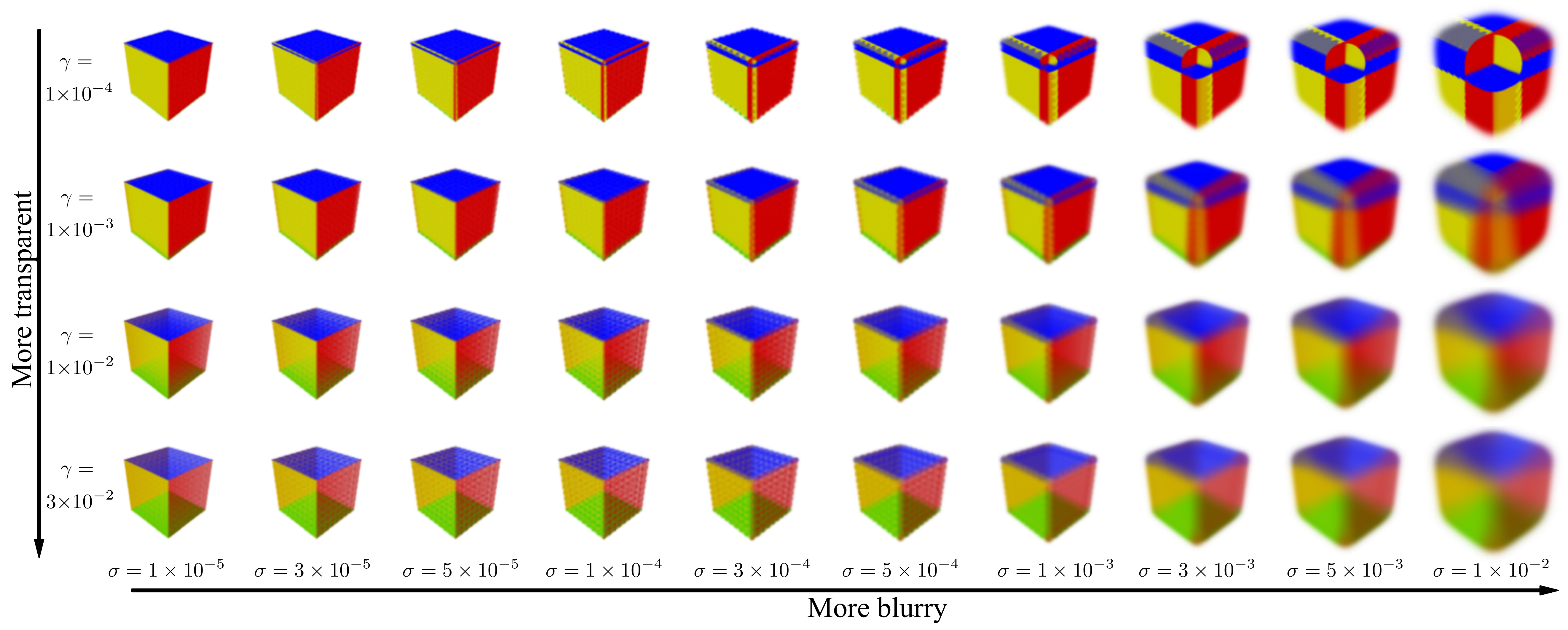}
    \caption{Different rendering effects achieved by our proposed SoftRas renderer. We show how a colorized cube can be rendered in various ways by tuning the parameters of SoftRas. In particular, by increasing $\gamma$, SoftRas can render the object with more tranparency while more blurry renderings can be achieved via increasing $\sigma$. As $\gamma \rightarrow 0$ and $\sigma \rightarrow 0$, one can achieve rendering effect closer to standard rendering.}
    \label{fig:app_supple_sigma_gamma}
\end{figure*}
	\section{Gradient Computation}
\label{sec:suppl_gradient}


In this section, we provide more analysis on the variants of the probability representation (Section~\ref{sec:prob_map}) and aggregate function (Section~\ref{sec:aggregate}), in terms of the mathematical formulation and the resulting impact on the backward gradient. 

\subsection{Overview}
According to the computation graph in Figure~\ref{fig:computeGraph}, our gradient from rendered image $\img$ to vertices in mesh $\geometry$ is obtained by

\begin{equation}
	\equPartial{\img}{\geometry} = \equPartial{\img}{\uv}\equPartial{\uv}{\geometry} + \equPartial{\img}{\depth}\equPartial{\depth}{\geometry} + \equPartial{\img}{\normal}\equPartial{\normal}{\geometry}.
\end{equation}

\noindent While $\equPartial{\uv}{\geometry}, \equPartial{\depth}{\geometry}, \equPartial{\img}{\normal}$ and $\equPartial{\normal}{\geometry}$ can be easily obtained by inverting the projection matrix and the illumination models, $\equPartial{\img}{\uv}$ and $\equPartial{\img}{\depth}$ do not exist in conventional rendering pipelines.
Our framework introduces an intermediate representation, probability map $\Dmap$, that factorizes the gradient $\equPartial{\img}{\uv}$ to $\equPartial{\img}{\Dmap} \equPartial{\Dmap}{\uv}$, enabling the differentiability of $\equPartial{\img}{\uv}$.
Further, we obtain $\equPartial{\img}{\depth}$ via the proposed aggregate function.
In the following context, we will first address the gradient $\equPartial{\Dmap}{\uv}$ in Section~\ref{sec:suppl_prob} and gradient $\equPartial{\img}{\Dmap}$ and  $\equPartial{\img}{\depth}$ in Section~\ref{sec:suppl_aggr}.

\subsection{Probability Map Computation}
\label{sec:suppl_prob}

The probability maps $\{\equDis_j^i\}$ based on the relative position between a given triangle $\face_j$ and pixel $\pixel_i$ are obtained via \textit{sigmoid function} with temperature $\equSigma$ and distance metric $\equDMetric(i, j)$:

\begin{equation}
	\equDis_j^i = \frac{1}{1 + \exp\left(-\frac{\equDMetric(i, j)}{\equSigma}\right)},
\end{equation}

\noindent where the metric $\equDMetric$ essentially satisfies: (1) $\equDMetric(i, j) > 0$ if $\pixel_i$ lies inside $\face_j$; (2) $\equDMetric(i, j) < 0$ if $\pixel_i$ lies outside $\face_j$ and (3) $\equDMetric(i, j) = 0$ if $\pixel_i$ lies exactly on the boundary of $\face_j$. The positive scalar $\equSigma$ controls the sharpness of the probability, where $\equDis_j$ converges to a binary mask as $\equSigma \rightarrow 0$.

We introduce two candidate metrics, namely \textit{signed Euclidean distance}  and \textit{barycentric metric}. We represent $\pixel_i$ using \textit{barycentric coordinate} $\equBaryMat_{j}^i \in \mathbb{R}^{3}$ defined by $\face_j$:

\begin{equation}
	\equBaryMat_{j}^i = \equFace_j^{-1} \equPixel_i,
\end{equation}

\noindent where $\equFace_j = \begin{bmatrix}
	\equX_1 & \equX_2 & \equX_3 \\
	\equY_1 & \equY_2 & \equY_3 \\
	1 & 1 & 1 \\
	\end{bmatrix}_{\face_j}$ and $\equPixel_i = \begin{bmatrix}
x \\
y \\
1
\end{bmatrix}_{\pixel_i}$.

\subsubsection{Euclidean Distance}

Let $\equBaryTMat_{j}^i \in \mathbb{R}^{3}$ be the barycentric coordinate of the point on the edge of $\face_j$ that is closest to $\pixel_i$. The signed Euclidean distance $\equDMetric_E(i, j)$ from $\pixel_i$ to the edges of $\face_j$ can be computed as:

\begin{align}
	\equDMetric_E(i, j) &= \equSig_{j}^i \left \lVert \equFace_j	(\equBaryTMat_{j}^i - \equBaryMat_{j}^i) \right \rVert_2^2 \nonumber \\ 
	&=  \equSig_{j}^i \left \lVert \equFace_j \equBaryTMat_{j}^i - \equPixel_i \right \rVert_2^2, 
\end{align}

\noindent where $\sig_j^i$ is a sign indicator defined as  $\sig_j^i = \{+1, \mathrm{if} \ \pixel_i \in \face_j; -1, \mathrm{otherwise}\}$.
%

Then the partial gradient $\equPartial{\equDMetric_E(i, j)}{\equFace_j}$ can be obtained via:

\begin{equation}
\equPartial{\equDMetric_E(i, j)}{\equFace_j} = 2\equSig_{j}^i\left(\equFace_j \equBaryTMat_{j}^i - \equPixel_i \right) \left(\equBaryTMat_{j}^i\right)^T.
\end{equation}

\begin{figure*}[h!]
  	\centering
	\includegraphics[width=1.0\linewidth]{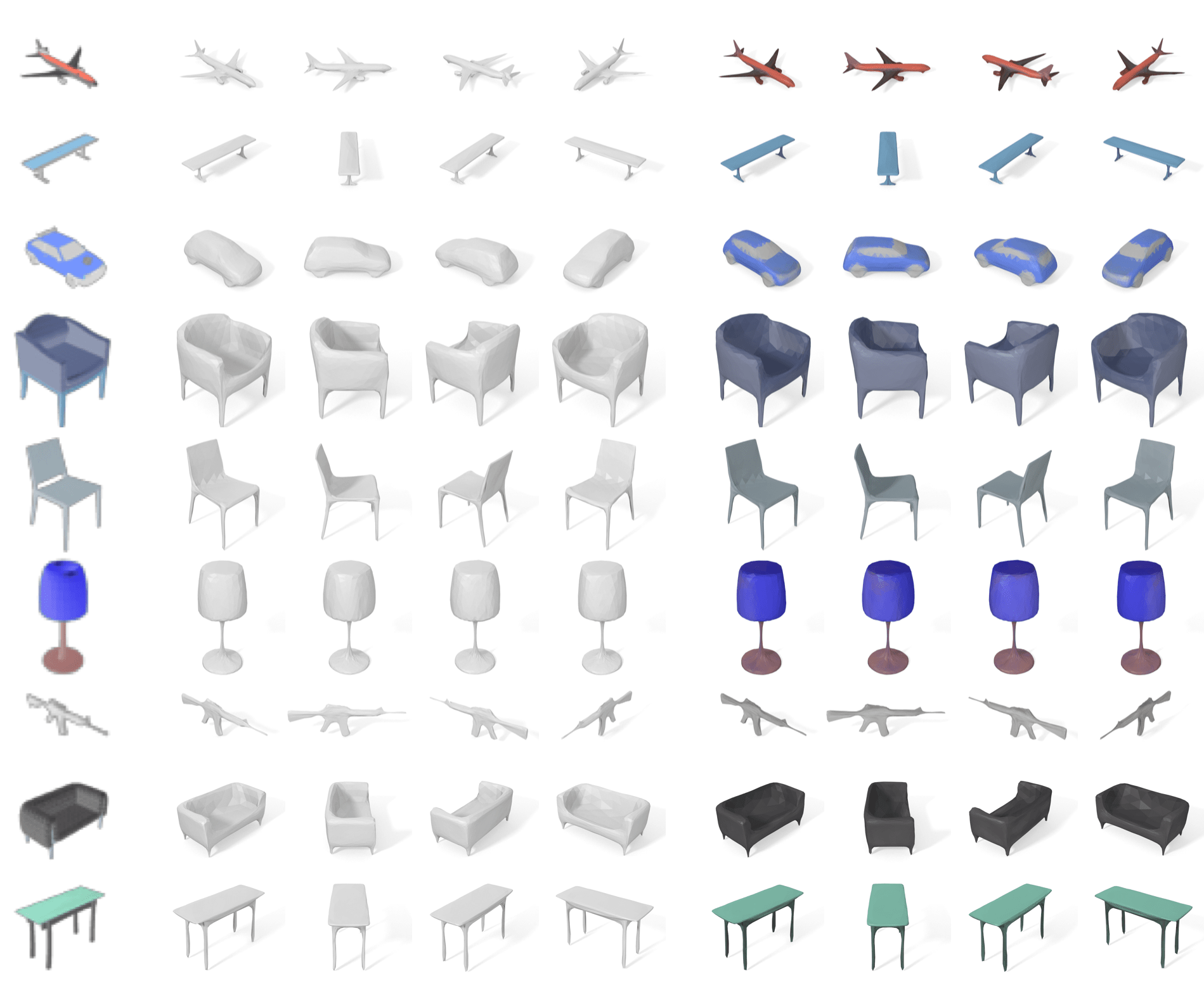}
	\caption{More single-view reconstruction results. Left: input image; middle: reconstructed geometry; right: colorized reconstruction.}
	\label{fig:supple_single_view_recon_more}
\end{figure*}

\subsubsection{Barycentric Metric}

We define the barycentric metric $\equDMetric_B(i, j)$ as the minimum of barycentric coordinate:

\begin{equation}
	\equDMetric_B(i, j) = \min\{\equBaryMat_{j}^i\}
\end{equation}

\noindent let $s =  \underset{k}\argminE \, (\equBaryMat_{j}^i)^{(k)}$, then the gradient from $\equDMetric_B(i, j)$ to $\equFace_j$ can be obtained through:

\begin{align}
	\equPartial{\equDMetric_B(i, j)}{\left(\equFace_j\right)^{(k,l)}} &= \equPartial{\min\{\equBaryMat_{j}^i\}}{\left(\equFace_j\right)^{(k,l)}}\nonumber\\ 
	&= \equPartial{\left(\equBaryMat_{j}^i\right)^{(s)}}{\equFace_j^{-1}} \equPartial{\equFace_j^{-1}}{\left(\equFace_j\right)^{(k,l)}} \nonumber\\ 
	&= -\sum_t \left(\equPixel_i\right)^{(t)} \left(\equFace_j^{-1}\right)^{(s,k)}\left(\equFace_j^{-1}\right)^{(l,t)},
\end{align}

\noindent where $k$ and $l$ are the indices of $\equFace_j$'s element.



\subsection{Aggregate function}
\label{sec:suppl_aggr}

\subsubsection{Softmax-based Aggregate Function}

According to $\aggregate_S(\cdot)$, the output color is:

\begin{equation}
\equImage^i = \aggregate_S(\{\equColor_j^i\}) = \sum_j \equSum_j^i \equColor_j^i + \equSum_b^i \equColor_b,
\end{equation}

\noindent where the weight $\{\equSum_j\}$ is obtained based on the relative depth $\{\equZ_j\}$ and the screen-space position of triangle $\face_j$ and pixel $\pixel_i$ as indicated in the following equation:

\begin{equation}
\equSum_j^i = \frac{\equDis_j^i \exp{\left({\equZ_j^i}/{\equGamma}\right)}}{\sum_k \equDis_k^i \exp{\left({\equZ_k^i}/{\equGamma}\right)} + \exp\left(\epsilon/\equGamma\right)};
\end{equation}

\noindent $\equColor_b$ and $\equSum_b^i$ denote the color and weight of background respectively where

\begin{equation}
\equSum_b^i = \frac{\exp\left(\epsilon/{\equGamma}\right)}{\sum_k \equDis_k^i \exp{\left({\equZ_k^i}/{\equGamma}\right)} + \exp\left(\epsilon/\equGamma\right)};
\end{equation}

\noindent $\equZ_j^i$ is the clipped normalized depth. Note that we normalize the depth so that the closer triangle receives a larger $\equZ_j^i$ by

\begin{equation}
\equZ_j^i = \frac{\equZFar - Z_j^i}{\equZFar - \equZNear},
\end{equation}

\noindent where $Z_j^i$ denotes the actual clipped depth of $\face_j$ at $\pixel_i$, while $\equZNear$ and $\equZFar$ denote the far and near cut-off distances of the viewing frustum.

Specifically, the aggregate function $\aggregate_S(\cdot)$ satisfies the following three properties: (1) as $\equGamma \rightarrow 0$ and $\equSigma \rightarrow 0$, $\equSum^i$ converges to an one-hot vector where only the closest triangle contains the projection of $\pixel_i$ is one, which shows the consistency between $\aggregate_S(\cdot)$ and \textit{z-buffering}; (2) $\equSum_b^i$ is close to one only when there is no triangle that covers $\pixel_i$; (3) $\{\equSum_j^i\}$ is robust to z-axis translation. In addition, $\equGamma$ is a positive scalar that could balance out the scale change on z-axis.


The gradient $\equPartial{\equImage}{\equDis_j^i}$ and $\equPartial{\equImage}{\equZ_j^i}$ can be obtained as follows:


\begin{align}
\equPartial{\equImage^i}{\equDis_j^i} &= \sum_k \equPartial{\equImage^i}{\equSum_k^i} \equPartial{\equSum_k^i}{\equDis_j^i} + \equPartial{\equImage^i}{\equSum_b^i} \equPartial{\equSum_b^i}{\equDis_j^i} \nonumber\\
&= \sum_{k \neq j} -\equColor_k^i \frac{ \equSum_j^i \equSum_k^i}{\equDis_j^i} + \equColor_j^i(\frac{\equSum_j^i}{\equDis_j^i}-\frac{ \equSum_j^i \equSum_j^i}{\equDis_j^i}) - \equColor_b^i \frac{ \equSum_j^i \equSum_b^i}{\equDis_j^i} \nonumber\\
&= \frac{\equSum_j^i}{\equDis_j^i} (\equColor_j^i - \equImage^i)
\end{align}

\begin{align}
\equPartial{\equImage^i}{\equZ_j^i} &= \sum_k \equPartial{\equImage^i}{\equSum_k^i} \equPartial{\equSum_k^i}{\equZ_j^i} + \equPartial{\equImage^i}{\equSum_b^i} \equPartial{\equSum_b^i}{\equZ_j^i} \nonumber\\
&= \sum_{k \neq j} -\equColor_k^i \frac{ \equSum_j^i \equSum_k^i}{\equGamma} + \equColor_j^i(\frac{\equSum_j^i}{\equGamma}-\frac{ \equSum_j^i \equSum_j^i}{\equGamma}) -\equColor_b^i \frac{ \equSum_j^i \equSum_b^i}{\equGamma} \nonumber\\
&= \frac{\equSum_j^i}{\equGamma} (\equColor_j^i - \equImage^i)
\end{align}

\subsubsection{Occupancy Aggregate Function} Independent from color and illumination, the silhouette of the object can be simply described by an occupancy aggregate function $\aggregate_O(\cdot)$ as follows:

\begin{equation}
	\equImage_{sil}^i = \aggregate_O(\{\equDis_j^i\}) = 1 - \prod_j(1 - \equDis_j^i).
\end{equation}

Hence, the partial gradient $\equPartial{\equImage_{sil}^i}{\equDis_j^i}$ can be computed as follows:
\begin{equation}
	\equPartial{\equImage_{sil}^i}{\equDis_j^i}	 = \frac{1 - \equImage_{sil}^i}{1 - \equDis_j^i}.
\end{equation}


%


	\section{Forward Rendering Results}
\label{sec:suppl_forward}

As demonstrated in Figure~\ref{fig:app_supple_sigma_gamma}, our framework is able to directly render a given mesh, which cannot be achieved by any existing rasterization-based differentiable renderers~\cite{kato2018neural,loper2014opendr}.
In addition, compared to standard graphics renderer, SoftRas can achieve different rendering effects in a continuous manner thanks to its probabilistic formulation.
Specifically, by increasing $\sigma$, the key parameter that controls the sharpness of the screen-space probability distribution, we are able to generate more blurry rendering results. 
Furthermore, with increased $\gamma$, one can assign more weights to the triangles on the far end, naturally achieving more transparency in the rendered image.
As discussed in Section~\ref{sec:result_fitting} of the main paper, the blurring and transparent effects are the key for reshaping the energy landscape in order to avoid local minima.
	
\section{Network Structure}
\label{sec:suppl_network}

\begin{figure}[h!]
    \centering
    \includegraphics[width=0.1\textwidth]{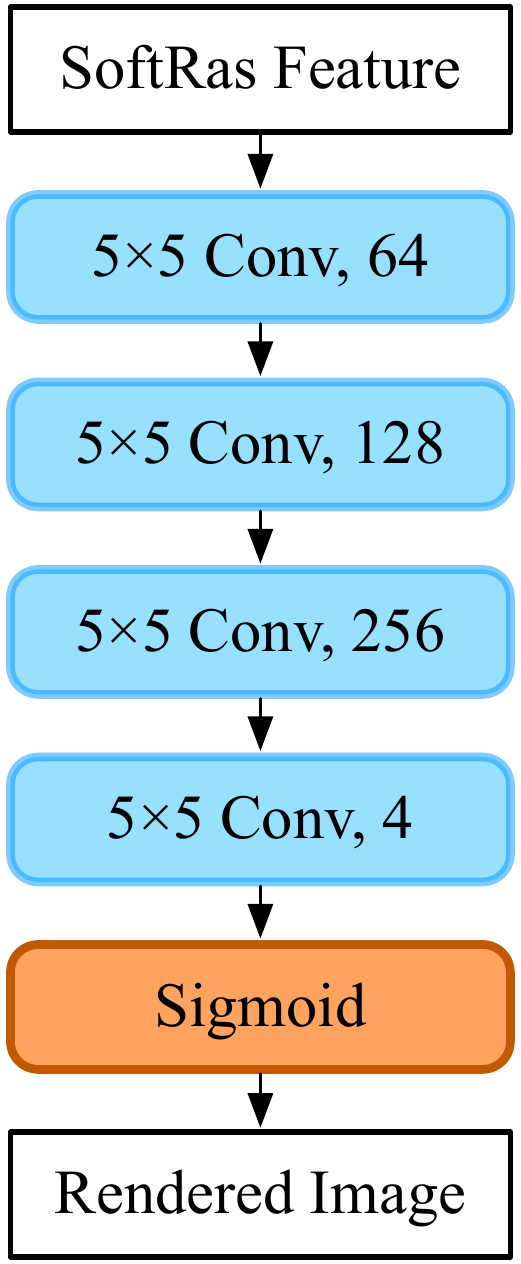}
    \caption{Network Architecture of $\aggregate_{N}$, an alternative color aggregate function that is implemented as a neural networks. 
    	}
    \label{fig:aggr_nn}
\end{figure}

We provide detailed structures for all neural networks that were mentioned in the main paper.
Figure~\ref{fig:aggr_nn} shows the structure of $\aggregate_{N}$ (Section~\ref{sec:aggregate} and \ref{sec:ablation}), an alternative color aggregate function that is implemented as a neural network.
In particular, input SoftRas features are first passed to four consecutive convolutional layers and then fed into a sigmoid layer to model non-linearity. 
We train $\aggregate_{N}$ with the output of a standard rendering pipeline as ground truth to achieve a \textit{parametric} differentiable renderer.

We employ an encoder-decoder architecture for our single-view mesh reconstruction.
The encoder is used as a feature extractor, whose network structure is shown in Figure~\ref{fig:supple_encoder}.
The detailed network structure of the color and shape generators are illustrated in Figure~\ref{fig:supple_decoder}(a) and (b) respectively.
Both networks (Figure~\ref{fig:framework}) share the same feature extractor.
The shape generators consists of three fully connected layers and outputs a per-vertex displacement vector that deforms a template mesh into a target model.
The color generator contains two fully connected streams: one for sampling the input image to build the color palette and the other one for selecting colors from the color palette to texture the sampling points.

\begin{figure}[tb!]
    \centering
    \includegraphics[width=0.1\textwidth]{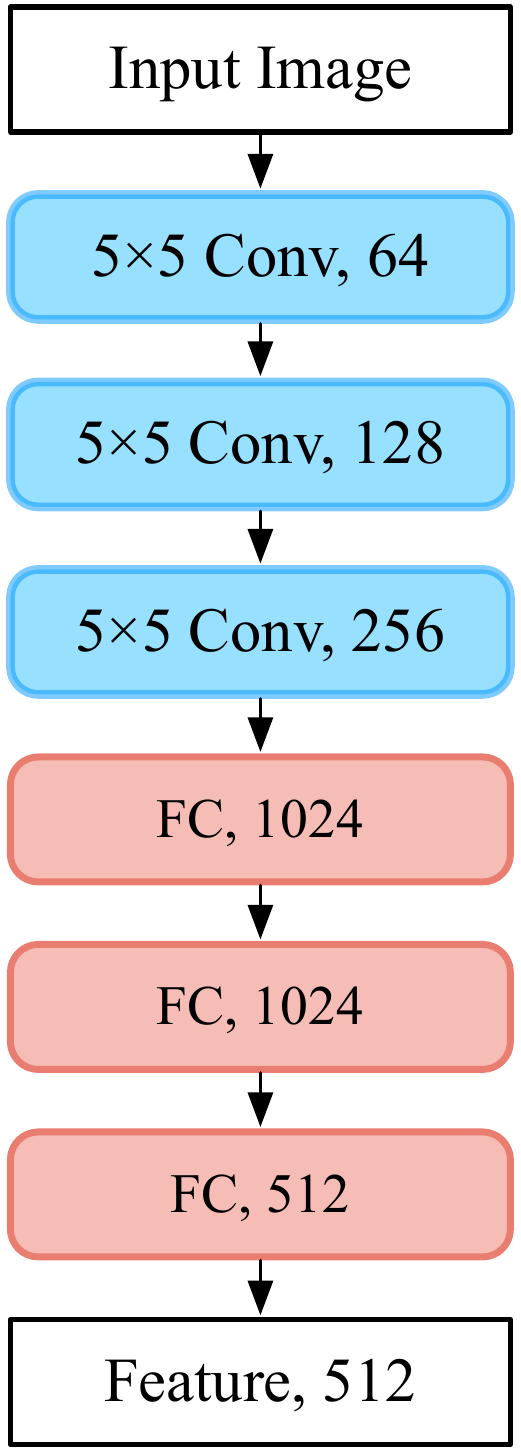}
    \caption{Network architecture of the feature extractor.}
    \label{fig:supple_encoder}
\end{figure}
\begin{figure}[tb!]
    \centering
    \includegraphics[width=0.36\textwidth]{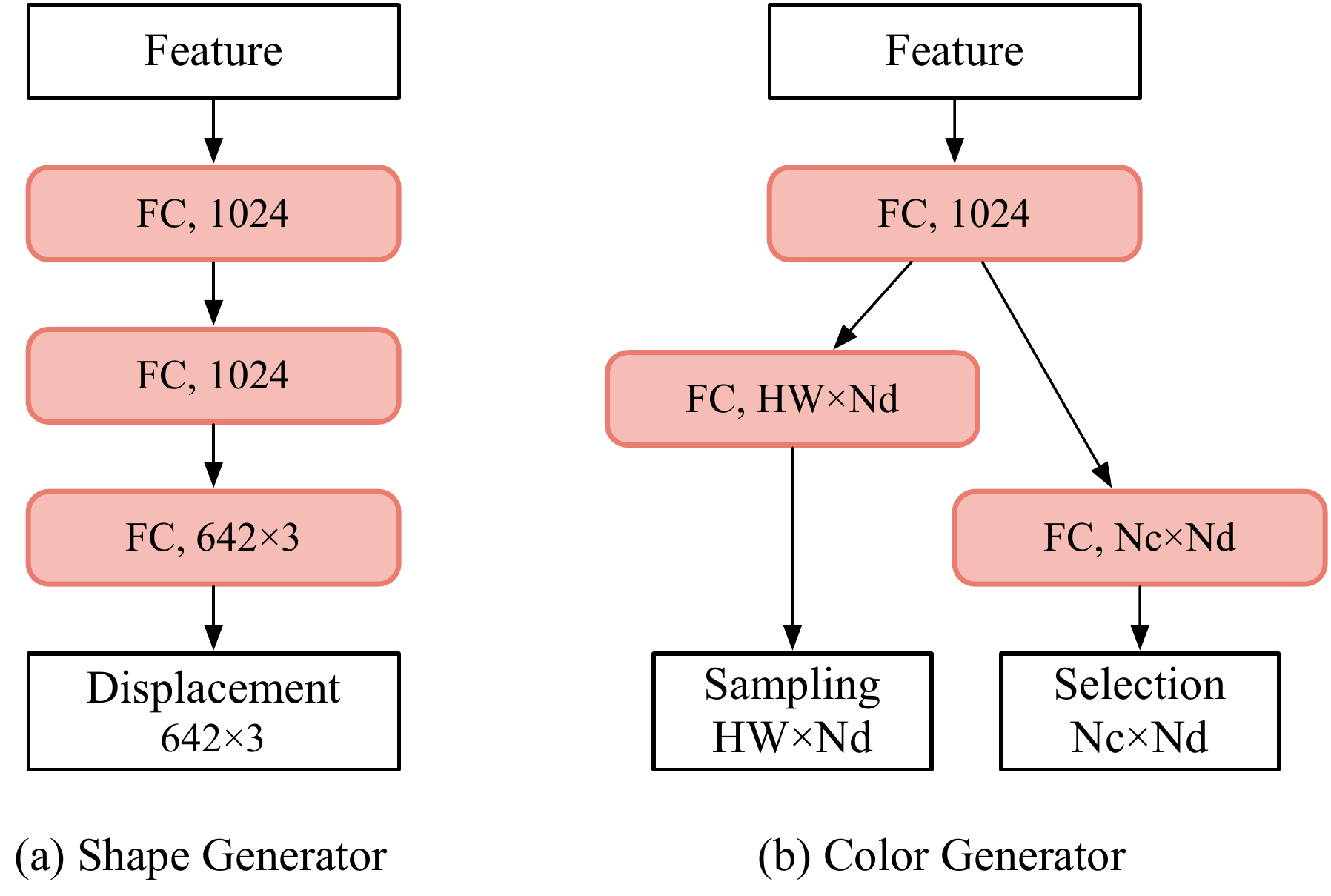}
    \caption{Network architectures of the shape and color generator.}
    \label{fig:supple_decoder}
\end{figure}

\section{More Results on Image-based 3D Reasoning}
\label{sec:suppl_more}

We show more results on single-view mesh reconstruction and image-base shape fitting.

\subsection{Single-view Mesh Reconstruction}
\label{sec:suppl_singe_view}



\subsubsection{Intermediate Mesh Deformation}

\begin{figure}[h]
  	\centering
	\includegraphics[width=1.0\linewidth]{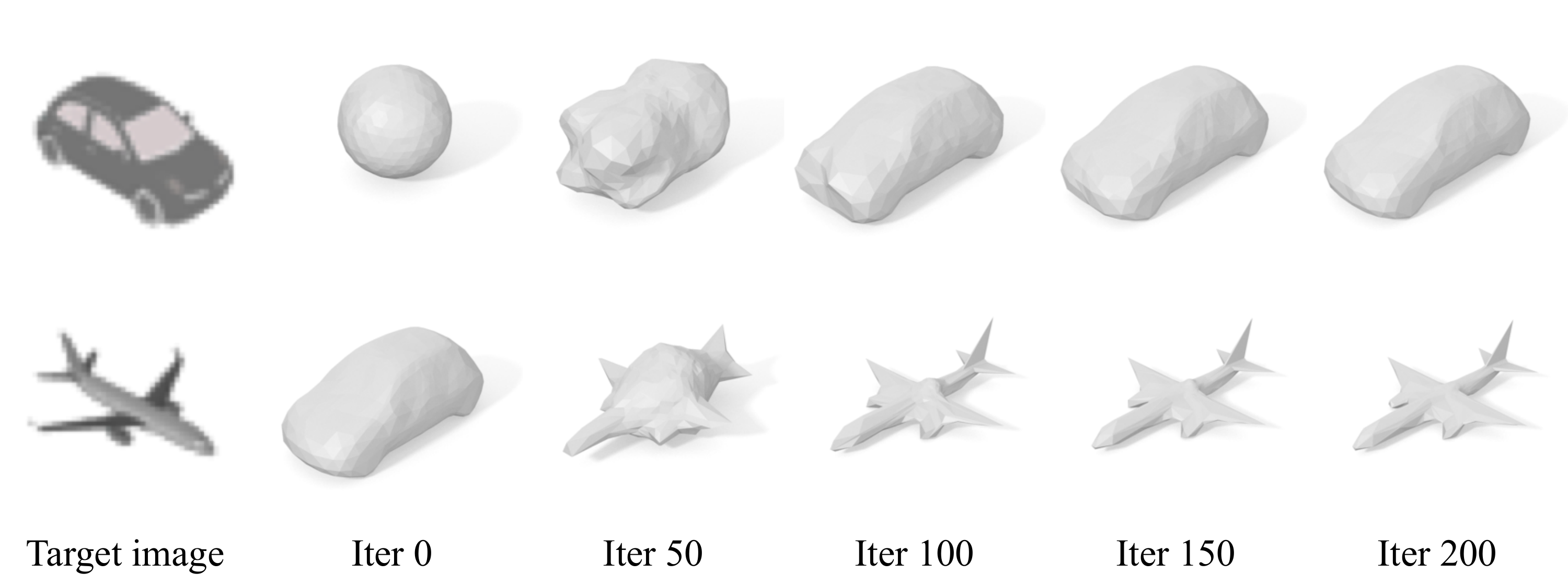}
	\caption{Visualization of intermediate mesh deformation during training. First row: the network deforms the input sphereto a desired car model that corresponds to the target image.  Second row:  the generated car model is further deformed to reconstruct the airplane.}
	\label{fig:supple_interpolation}
\end{figure}

\begin{figure}[h]
  	\centering
	\includegraphics[width=\linewidth]{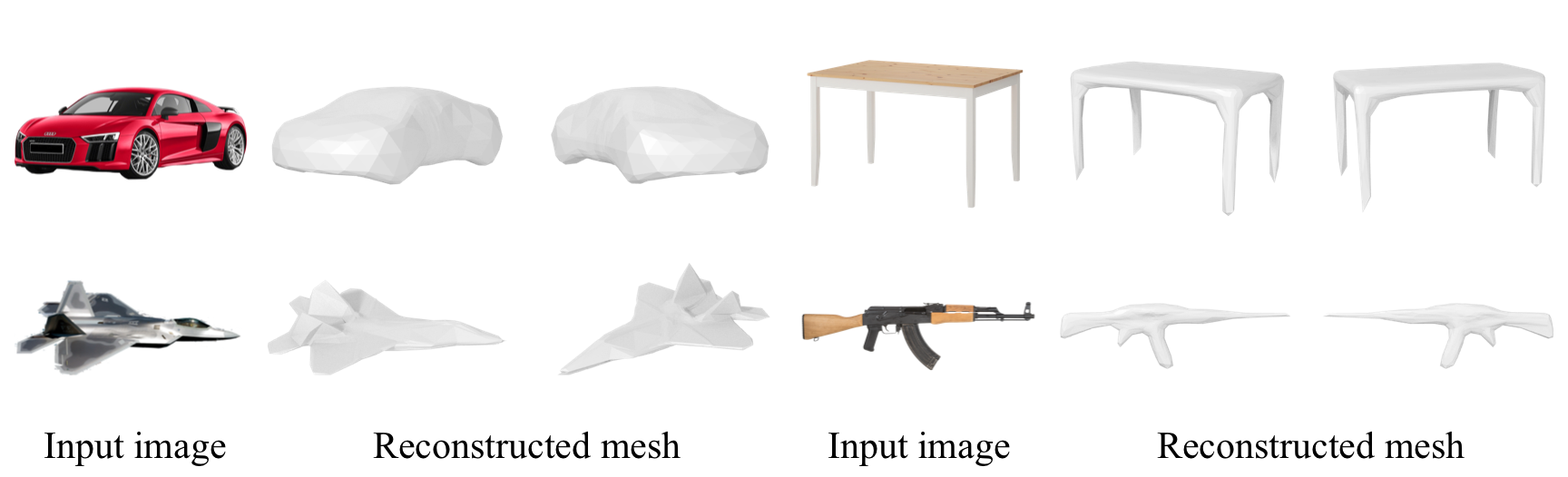}
	\captionof{figure}{Single-view reconstruction results on real images.}
	\label{fig:supple_recon_real_image}
\end{figure}

In Figure~\ref{fig:supple_interpolation}, we visualize the intermediate process of how an input mesh is deformed to a target shape after the supervision provided by SoftRas.
As shown in the first row, the mesh generator gradually deforms a sphere template to a desired car shape which matches the input image. 
We then change the target image to an airplane (Figure~\ref{fig:supple_interpolation} second row).
The network further deforms the generated car model to faithfully reconstruct the airplane.
In both examples, the mesh deformation can quickly converge to a high-fidelity reconstruction within 200 iterations, demonstrating the effectiveness of our SoftRas renderer.

\subsubsection{Single-view Reconstruction from Real Images}
We further evaluate our approach on real images.
As demonstrated in Figure~\ref{fig:supple_recon_real_image}, though only trained on synthetic data, our model generalizes well to real images and novel views with faithful reconstructions and fine-scale details, e.g. the tail fins of the fighter aircraft and thin structures in the rifle and table legs.

\subsubsection{More Reconstruction Results from ShapeNet}

We provide more reconstruction results in Figure~\ref{fig:supple_single_view_recon_more}.  
For each input image, we show its reconstructed geometry (middle) as well as the colored reconstruction (right).

\begin{figure}[h]
  	\centering
	\includegraphics[width=\linewidth]{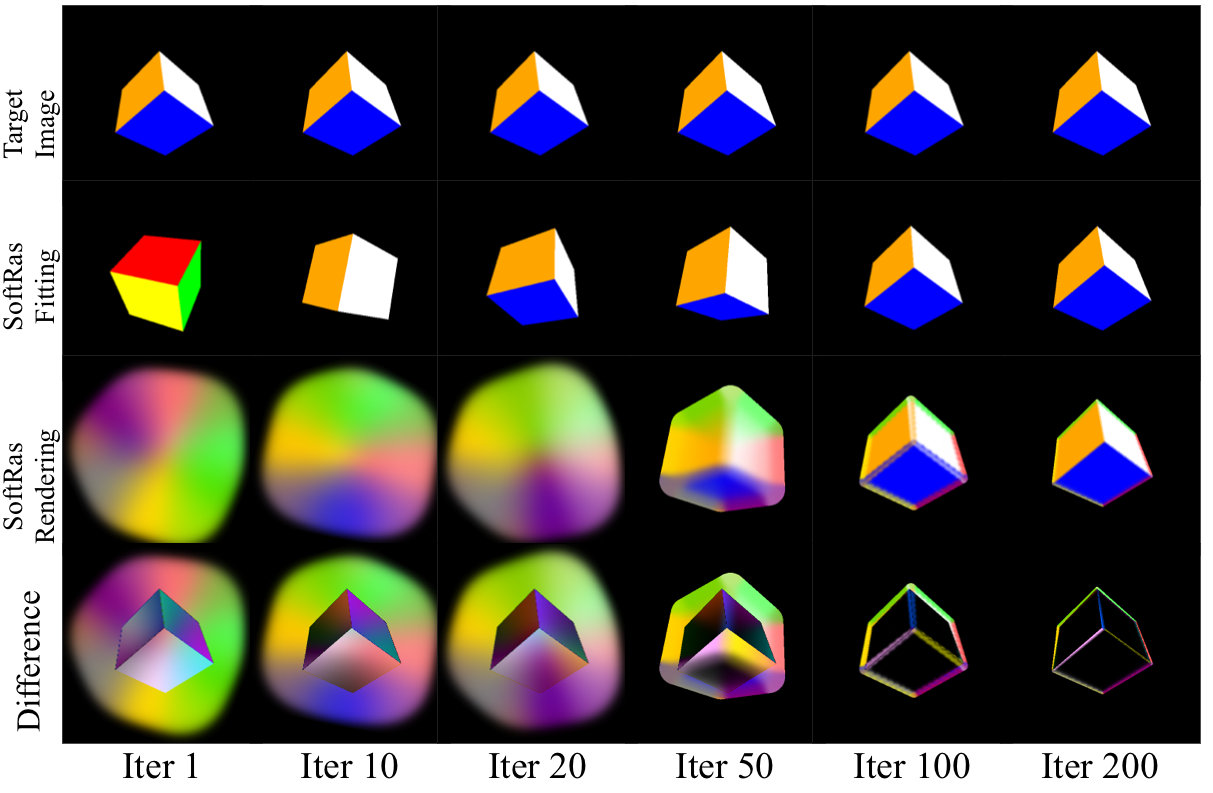}
	\caption{Intermediate process of fitting a color cube (second row) to a target pose shown in the input image (first row). The smoothened rendering (third row) that is used to escape local minimum, as well as the colorized fitting errors (fourth row), are also demonstrated.}
	\label{fig:supple_cube_nmr}
\end{figure}

\subsection{Fitting Process for Rigid Pose Estimation}
\label{sec:suppl_rigid_fit}

We demonstrate the intermediate process of how the proposed SoftRas renderer managed to fit the color cube to the target image in Figure~\ref{fig:supple_cube_nmr}.
Since the cube is largely occluded, directly leveraging a standard rendering is likely to lead to local minima (Figure~\ref{fig:cube}) that causes non-trivial challenges for any gradient-based optimizer.
By rendering the cube with stronger blurring at the earlier stage, our approach is able to avoid local minima, and gradually reduce the rendering loss until an accurate pose can be fitted.  


\subsection{Visualization of Non-rigid Body Fitting}
\label{sec:suppl_smpl_fit}

In Figure~\ref{fig:supple_smpl_softras}, we compare the intermediate processes of NMR~\cite{kato2018neural} and SoftRas during the task of fitting the SMPL model to the target pose. 
As the right hand of subject is completely occluded in the initial image, NMR fails to complete the task due to its incapability of flowing gradient to the occluded vertices.
In contrast, our approach is able to obtain the correct pose within 320 iterations thanks to the occlusion-aware technique.

%

\begin{figure}[H]
  	\centering
	\includegraphics[width=\linewidth]{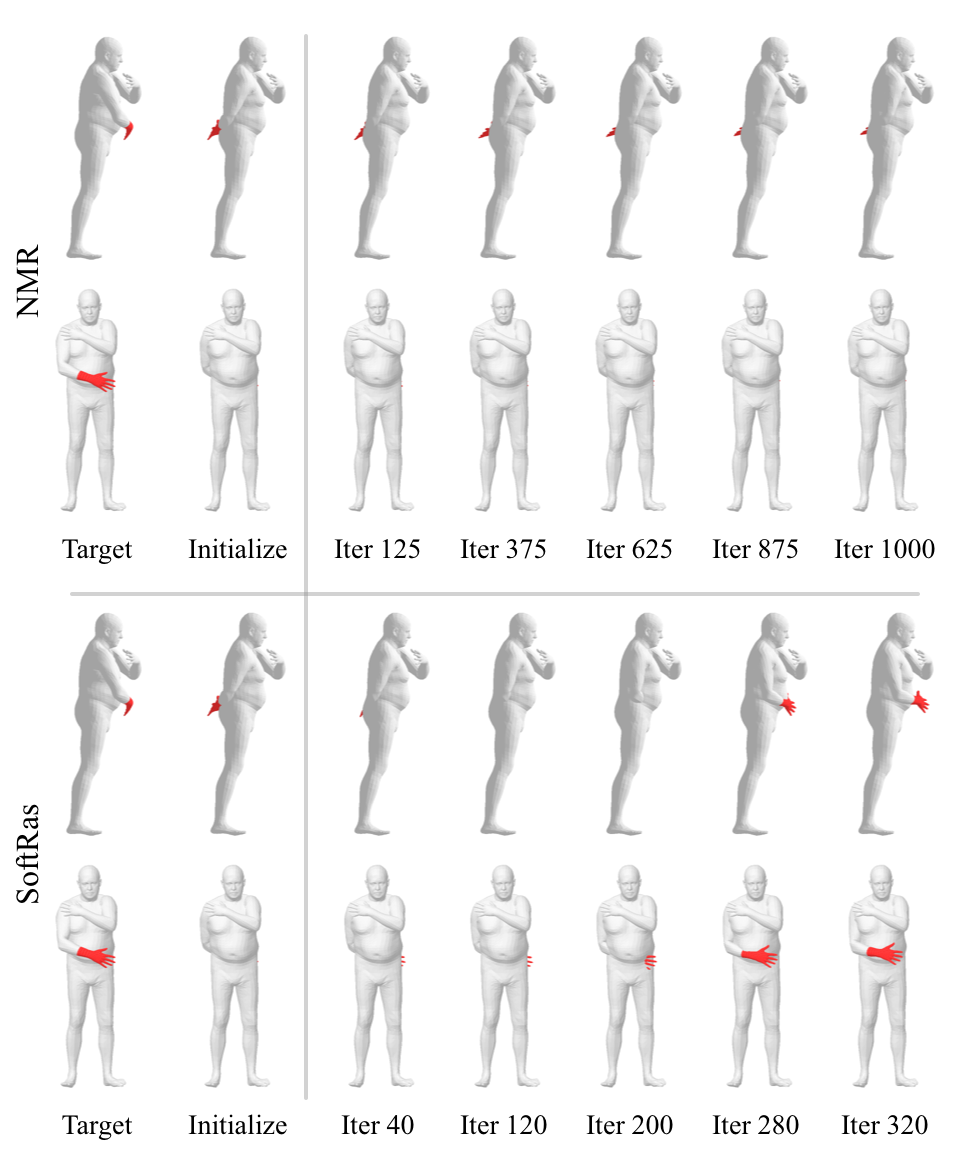}
	\caption{Comparisons of body shape fitting using NMR~\cite{kato2018neural} and our approach. Intermediate fitting processes of both methods are visualized.}
	\label{fig:supple_smpl_softras}
\end{figure}
}
{}

\ifthenelse{\equal{\final}{0}}
{
}
{}
\end{document}